\documentclass{article}
\PassOptionsToPackage{numbers,compress}{natbib}
\usepackage[preprint]{neurips_2023}
\usepackage[utf8]{inputenc} 
\usepackage[T1]{fontenc}    
\usepackage{hyperref}       
\usepackage{url}            
\usepackage{booktabs}       
\usepackage{amsfonts}       
\usepackage{nicefrac}       
\usepackage{microtype}      
\usepackage{xcolor}         
\usepackage[title]{appendix}

\usepackage{amsmath}
\usepackage{amssymb}
\usepackage{mathtools}
\usepackage{amsthm}
\usepackage{thmtools}
\usepackage{thm-restate}
\usepackage{floatrow}
\usepackage{sidecap}
\usepackage{subcaption}
\usepackage{microtype}
\usepackage{graphicx}
\usepackage{booktabs} 
\usepackage[english]{babel}
\usepackage{amsthm,amsmath,amsfonts,amssymb}
\usepackage{mathrsfs}
\usepackage[accsupp]{axessibility}  
\usepackage{enumitem}
\usepackage[percent]{overpic} 
\usepackage{bbm}
\usepackage{color,soul}
\usepackage{makecell}
\usepackage{tabularx,ragged2e}
\usepackage[framemethod=TikZ]{mdframed}
\usepackage{setspace}
\usepackage{array}
\newcolumntype{Y}{>{\centering\arraybackslash}X}
\newcolumntype{x}[1]{>{\centering}p{#1}}
\usepackage{comment}

\usepackage{todonotes}
\usepackage{xcolor}

\usepackage{xparse}
\usepackage{tikz}
\usepackage{calc}
\usetikzlibrary{calc}
\newcommand{\boxStart}[1]{\tikz[overlay,remember picture] \node (startPoint#1) {};}
\newcommand{\boxEnd}[1]{%
	\tikz[overlay,remember picture]{
		\node (endPoint#1) {};
		\draw[thin, gray, fill=yellow, fill opacity=0.3]
			($(startPoint#1)+(-0.3em,0.9em)$) rectangle
			($(endPoint#1)+(0.3em,-0.3em)$);}
}

\makeatletter
\newenvironment{lflalign}{%
  \def\align@preamble{%
    &\strut@                                                                        
    \setboxz@h{\@lign$\m@th\displaystyle{####}$}%
    \ifmeasuring@\savefieldlength@\fi                                               
    \set@field                                                                      
    \hfil                                                                           
    \tabskip\z@skip                                                                 
    &\setboxz@h{\@lign$\m@th\displaystyle{{}####}$}%
    \ifmeasuring@\savefieldlength@\fi                                               
    \set@field                                                                      
    \hfil                                                                           
    \tabskip\alignsep@                                                              
  }                                                                                 
  \flalign}
{\endflalign}
\makeatother

\definecolor{myDarkGray}{gray}{0.5}
\definecolor{myGray}{gray}{0.85}
\definecolor{myBlue}{RGB}{77,77,255}
\definecolor{myLightBlue}{RGB}{102,102,255}
\definecolor{myRed}{RGB}{255,26,26}
\definecolor{myMagenta}{RGB}{204,26,204}
\newcommand{\dataset}[1]{\textit{#1}}
\def\eqdef{\stackrel{\text{def}}{=}}
\newcommand{\crit}[1]{\textit{#1}}

\def\pY{\boldsymbol{\operatorname{p}}_\textsc{Y}\mkern-1.2mu}

\def\bx{\boldsymbol{x}}
\def\by{\boldsymbol{y}}
\def\bp{\boldsymbol{p}}
\def\bpp{\widetilde{\boldsymbol{p}}}
\def\bpi{{p_i}}
\def\bpj{{p_j}}
\def\bppi{{\widetilde{p}_i}}
\def\btheta{{\boldsymbol{\theta}}}
\def\expectation{\operatorname{\mathbb{E}}}
\newcommand{\entropy}[1]{\operatorname{\mathbb{H}}\mkern-0.4mu({#1})}
\newcommand{\entropyBig}[1]{\operatorname{\mathbb{H}}\mkern-2.9mu\big({#1}\big)}
\renewcommand{\paragraph}[1]{\noindent{\normalsize\bfseries #1}\hspace{.05em}} 

\theoremstyle{plain}
\newtheorem{theorem}{Theorem}[section]

\theoremstyle{definition}

\theoremstyle{remark}

\title{Selective Mixup Helps with Distribution Shifts,\\But Not (Only) because of Mixup}

\author{%
  Damien Teney \\
  Idiap Research Institute\\
  Martigny, Switzerland \\
  \texttt{damien.teney@idiap.ch} \\
  \And
  Jindong Wang \\
  Microsoft Research Asia \\
  Beijing, China \\
  \texttt{jindong.wang@microsoft.com} \\
  \And
  Ehsan Abbasnejad \\
  AIML, University of Adelaide \\
  Adelaide, Australia \\
  \texttt{ehsan.abbasnejad@adelaide.edu.au} \\
}

\begin{document}

\maketitle

\begin{abstract}
\textbf{Context.}
Mixup is a highly successful technique to improve generalization of neural networks by augmenting the training data with combinations of random pairs. Selective mixup is a family of methods that apply mixup to specific pairs, e.g.~only combining examples across classes or domains. These methods have claimed remarkable improvements on benchmarks with distribution shifts, but their mechanisms and limitations remain poorly understood.
 
\vspace{4pt}\textbf{Findings.}
We examine an overlooked aspect of selective mixup that explains its success in a completely new light. We find that the non-random selection of pairs affects the training distribution and improve generalization by means completely unrelated to the mixing. For example in binary classification, mixup across classes implicitly resamples the data for a uniform class distribution ---~a classical solution to label shift. We show empirically that this implicit resampling explains much of the improvements in prior work. Theoretically, these results rely on a ``regression toward the mean'', an accidental property that we identify in several datasets.

\vspace{4pt}\textbf{Takeaways.}
We have found a new equivalence between two successful methods: selective mixup and resampling. We identify limits of the former, confirm the effectiveness of the latter, and find better combinations of their respective benefits.
\end{abstract}

\section{Introduction}

Mixup and its variants are some of the few methods that improve generalization across tasks and modalities with no domain-specific information~\cite{mixup}.
Standard mixup replaces training data with linear combinations of random pairs of examples, proving successful e.g.\ for image classification~\cite{yun2019cutmix}, semantic segmentation~\cite{islam2023segmix}, natural language processing~\cite{manifold_mix}, and speech processing~\cite{meng2021mixspeech}.

\begin{figure*}[t!]
  \includegraphics[width=.8\textwidth]{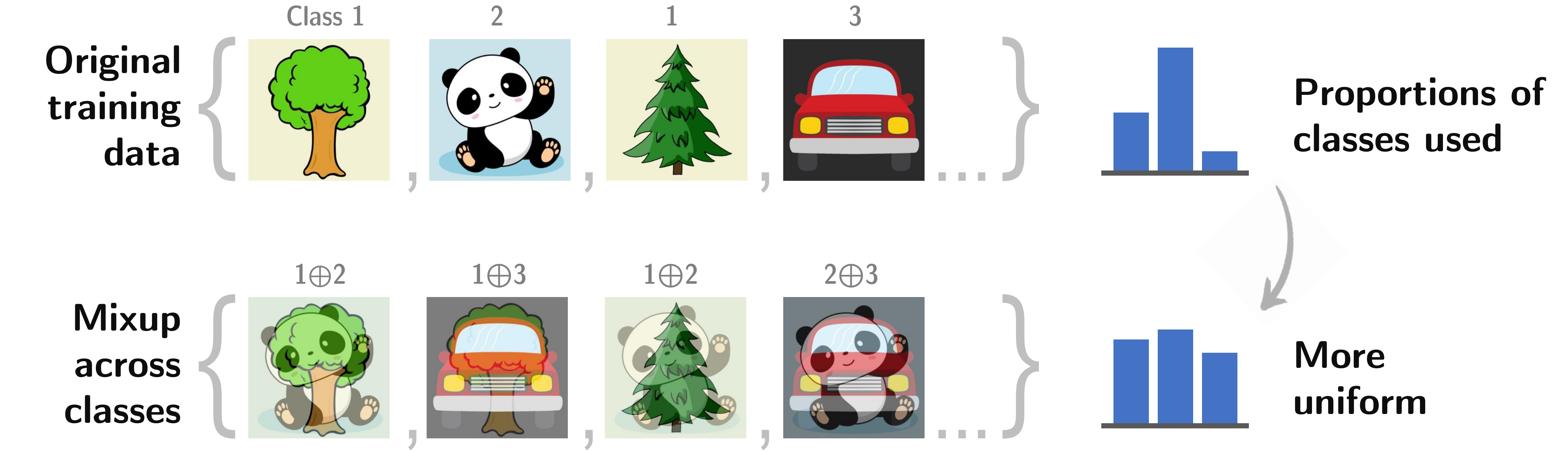}
  \caption{Selective mixup is a family of methods that replace the training data with combined pairs of examples fulfilling a predefined criterion, e.g.\ pairs of different classes. 
  As an overlooked side effect, this modifies the training distribution: in this case, sampling classes more uniformly. This effect is responsible for much of the resulting improvements in OOD generalization.\label{figTeaser}}
\end{figure*}

This paper focuses on scenarios of distribution shift
and on variants of mixup that improve out-of-distribution (OOD) generalization.
We examine the family of methods that apply mixup on selected pairs of examples, which we refer to as \emph{selective mixup}~\cite{hwang2022selecmix,li2023data,lu2022semantic,palakkadavath2022improving,tian2023cifair,xu2020adversarial,yao2022improving}.
Each of these method uses a predefined criterion.%
\footnote{We focus on the basic implementation of selective mixup as described by~\citet{yao2022improving}, i.e.\ without additional regularizers or modifications of the learning objective described in various other papers.}
For example, some methods combine examples across classes~\cite{yao2022improving} (Figure~\ref{figTeaser}) or across domains~\cite{xu2020adversarial,li2023data,lu2022semantic}.
These simple heuristics have claimed remarkable improvements on benchmarks
such as DomainBed~\cite{gulrajani2020search}, WILDS~\cite{koh2021wilds}, and Wild-Time~\cite{yao2022wild}.

Despite impressive empirical performance, the theoretical mechanisms of selective mixup remain obscure.
For example, the selection criteria proposed in~\cite{yao2022improving} include the selection of pairs of the same class\,/\,different domains, but also the exact opposite. This raises questions:
\setlist{nolistsep,leftmargin=*}
\begin{enumerate}[topsep=-2pt,itemsep=2pt,labelindent=*,labelsep=10pt,leftmargin=23pt]
\item What mechanisms are responsible for the improvements of selective mixup?
\item What makes each selection criterion suitable to any specific dataset?
\end{enumerate}

This paper presents surprising answers by highlighting an overlooked side effect of selective mixup.
\textbf{The non-random selection of pairs implicitly biases the training distribution
and improve generalization by means completely unrelated to the mixing}.
We observe empirically that simply 
forming mini-batches with all instances of the selected pairs (without mixing them)
often produces the same improvements as mixing them.
This critical ablation was absent from prior studies.

We also analyze theoretically the resampling induced by different selection criteria.
We find that conditioning on a ``different attribute'' (e.g.\ combining examples across classes or domains)
brings the training distribution of this attribute closer to a uniform one.
Consequently, the imbalances in the data often ``regress toward the mean'' with selective mixup.
We verify empirically that several datasets do indeed shift toward a uniform class distribution in their test split (see Figure~\ref{figRegression}).
We also find remarkable correlation between improvements in performance and the reduction in divergence of training/test distributions due to selective mixup.
This also predicts an new failure mode of selective mixup when the above property does not hold (see Appendix~\ref{appendixPredictedFailure}).

\setlength{\textfloatsep}{10pt} 
\floatsetup[figure]{margins=centering,capposition=bottom,floatwidth=\textwidth}

\paragraph{Our contributions are summarized as follows.}
\setlist{nolistsep,leftmargin=*}
\begin{itemize}[topsep=-2pt,itemsep=4pt,labelindent=2pt,labelsep=*,leftmargin=12pt]
    \item
    We point out an overlooked resampling effect when applying selective mixup (Section~\ref{secBias}).
    
    \item
    We show theoretically that certain selection criteria induce a bias in the distribution of features and/or classes equivalent to a ``regression toward the mean'' (Theorem~\ref{thmRegression}).
    In binary classification for example, selecting pairs across classes is equivalent to sampling uniformly over classes, the standard approach to address label shift and imbalanced data.

    \item We verify empirically that multiple datasets indeed contain a regression toward a uniform class distribution across training and test splits (Section~\ref{secExperimentsRegression}).
    We also find that improvements from selective mixup correlate with reductions in divergence of training/test distributions over labels and/or covariates.
    This strongly suggests that resampling is the main driver for these improvements.

    \item
    We compare many selection criteria and resampling baselines on five datasets.
    In all cases, improvements with selective mixup are partly or fully explained by resampling effects (Section~\ref{secExperiments}).
\end{itemize}

\paragraph{The implications for future research are summarized as follows.}
\setlist{nolistsep,leftmargin=*}
\begin{itemize}[topsep=-2pt,itemsep=4pt,labelindent=2pt,labelsep=*,leftmargin=12pt]
    \item
We connect two areas of the literature by showing that
{selective mixup is sometimes equivalent to resampling}, a classical strategy for distribution shifts~\cite{garg2023rlsbench,idrissi2022simple}.
This hints at possible benefits from advanced methods for label shift and domain adaptation on benchmarks with distribution shifts.

    \item The resampling explains why different criteria in selective mixup benefit different datasets: they
    affect the distribution of features and/or labels 
    and therefore address covariate and/or label shift.

    \item There is a risk of overfitting to the benchmarks: we show that much of the observed improvements rely on the accidental property of a ``regression toward the mean'' in the datasets examined.
\end{itemize}

\section{Background: mixup and selective mixup}

\paragraph{Notations.}
We consider a classification model $f_\btheta : \mathbb{R}^d \rightarrow [0,1]^C$ of learned parameters $\btheta$.
It maps an input vector $\bx \in \mathbb{R}^d$ to a vector $\by$ of scores over $C$ classes.
The training data for such a model is typically a set of labeled examples $\mathcal{D} = \{(\bx_i,\by_i,d_i)\}_{i=1}^{n}$
where $\by_i$ are one-hot vectors encoding ground-truth labels, and $d_i \in \mathbb{N}$ are optional discrete domain indices.
Domain labels are sometimes available e.g.\ in datasets with different image styles~\cite{li2017deeper} or collected over different time periods~\cite{koh2021wilds}.

\paragraph{Training with ERM.}
Standard empirical risk minimization (ERM) optimizes the model's parameters
for $\operatorname{min}_\btheta \,\mathcal{R}(f_\btheta, \mathcal{D})$
where the expected training risk, for a chosen loss function $\mathcal{L}$, is defined as:
\abovedisplayskip=4pt
\belowdisplayskip=4pt
\begin{align}\label{eqErm}
    \mathcal{R}(f_\btheta, \mathcal{D}) \;=\;
    \expectation_{(\bx, \by) \in \mathcal{D}} \, \mathcal{L}\big(f_\btheta(\bx), \by\big).
\end{align}
An empirical estimate is obtained with an arithmetic mean over instances of the dataset $\mathcal{D}$.

\paragraph{Training with mixup.}
Standard mixup essentially replaces training examples with linear combinations of random pairs in both input and label space.
We formalize it by redefining the training risk.\vspace{-11pt}
\abovedisplayskip=4pt
\belowdisplayskip=4pt
\begin{align}\label{eqMixup}
    \mathcal{R}_\textrm{mixup}(f_\btheta, \mathcal{D}) \;=\;
    \expectation_{(\bx, \by) \in \mathcal{D}} \,
    \mathcal{L}\big(f( c\,\bx \!+\! (1\!-\!c) \widetilde{\bx}, \;c\,\by \!+\! (1\!-\!c)\,\widetilde{\by})\big)\\
    \textrm{with mixing coefficients}~ c \sim \mathcal{B}(2,2)~~~
    \textrm{and paired examples}~ (\widetilde{\bx}, \widetilde{\by}) \sim \mathcal{D}.
\end{align}
The expectation is approximated by sampling different coefficients and pairs at every training iteration.

\paragraph{Selective mixup.}
While standard mixup combines random pairs,
selective mixup only combines pairs that fulfill a predefined criterion.
To select these pairs, the method starts with the original data $\mathcal{D}$, 
then for every
$({\bx},{\by},{d}) \in \mathcal{D}$
it selects a
$(\widetilde{\bx},\widetilde{\by},\widetilde{d}) \in \mathcal{D}$
such that they fulfill the criterion represented by the predicate
$\operatorname{Paired}\!\big(\cdot, \cdot)$.
For example, the criterion \crit{same class\,/\,different domain} a.k.a.\ ``{intra-label LISA}'' in \cite{yao2022improving} is implemented as follows:
\abovedisplayskip=2pt
\belowdisplayskip=0pt
\setlength{\jot}{2pt}
\begin{subequations}
\begin{lflalign}\label{eqPredicateExamples}
    \operatorname{Paired}\!\big( (\bx_i, \by_i, d_i), (\widetilde{\bx}_i, \widetilde{\by}_i, \widetilde{d}_i) \big)\!=\!\textit{true}
    ~~\textrm{iff}~~
    (\widetilde{\by} \!=\! \by) \land (\widetilde{d} \!\neq\! d)~~
    &\textit{(same class\,/\,diff. domain)}\\
    \hspace{-12pt}\textrm{Other examples:}\nonumber\\
    \operatorname{Paired}\!\big( (\bx_i, \by_i, d_i), (\widetilde{\bx}_i, \widetilde{\by}_i, \widetilde{d}_i) \big)\!=\!\textit{true}
    ~~\textrm{iff}~~
    (\widetilde{\by} \!\neq\! \by)
    &\textit{(different class)}\\
    \label{eqPredicateExamplesLast}
    \operatorname{Paired}\!\big( (\bx_i, \by_i, d_i), (\widetilde{\bx}_i, \widetilde{\by}_i, \widetilde{d}_i) \big)\!=\!\textit{true}
    ~~\textrm{iff}~~
    (\widetilde{d} \!=\! d)
    &\textit{(same domain)}
\end{lflalign}
\end{subequations}

\section{Selective mixup modifies the training distribution}
\label{secBias}

The new claims of this paper comprise two parts.
\setlist{nolistsep,leftmargin=*}
\begin{enumerate}[topsep=-2pt,itemsep=4pt,labelindent=0pt,labelsep=*,leftmargin=12pt]
\item Estimating the training risk with selective mixup (Eq.~\ref{eqMixup}) uses a different sampling of examples from $\mathcal{D}$ than ERM~(Eq.~\ref{eqErm}). We demonstrate this theoretically in this section.

\item We hypothesize that this different sampling of training examples influences the
generalization properties of the
learned model, regardless of the mixing operation.
We verify this empirically in Section~\ref{secExperiments} using ablations of selective mixup that omit the mixing operation~---~a critical baseline absent from prior studies.
\end{enumerate}

\paragraph{Training distribution.}
This distribution refers to the examples sampled from $\mathcal{D}$ to estimate the training risk (Eq.~\ref{eqErm} or \ref{eqMixup}) ~---~whether these are then mixed or not.
The following discussion focuses on distributions over classes~($\by$) but analogous arguments apply to covariates~($\bx$) and domains~($d$).

\paragraph{With ERM}, the training distribution equals the dataset distribution
because the expectation in Eq.~(\ref{eqErm}) is over uniform samples of $\mathcal{D}$.
We obtain an empirical estimate
by averaging all one-hot labels, giving the vector of discrete probabilities
$\pY(\mathcal{D}) \,=\, \oplus_{(\bx,\by) \in \mathcal{D}}\,\by\,/\,|\mathcal{D}|$
where $\oplus$~is the element-wise sum.

\paragraph{With selective mixup}, evaluating the risk (Eq.~\ref{eqMixup})
requires pairs of samples. 
The first element of a pair is sampled uniformly, yielding the same $\pY(\mathcal{D})$ as ERM.
The second element is selected as described above, using the first element and one chosen predicate $\operatorname{Paired}(\cdot, \cdot)$ e.g.\ from (\ref{eqPredicateExamples}--\ref{eqPredicateExamplesLast}).
For our analysis, we denote these ``second elements'' of the pairs as the virtual data:\\
\abovedisplayskip=-4pt
\belowdisplayskip=4pt
\begin{align}\label{eqVirtualData}
    \widetilde{\mathcal{D}} ~=~ \big\{
    (\widetilde{\bx}_i, \widetilde{\by}_i, \widetilde{d}_i) \sim \mathcal{D}:~~
    \operatorname{Paired}\!\big( (\bx_i, \by_i, d_i), (\widetilde{\bx}_i, \widetilde{\by}_i, \widetilde{d}_i) \big)\!=\!\operatorname{}{true},
    \,~~\forall \,i=1,\ldots,|\mathcal{D}|
    \big\}\,.
\end{align}
We can now analyze the overall training distribution of selective mixup.
An empirical estimate is
obtained by combining the distributions resulting from the two elements of the pairs, which gives the vector
$\pY(\mathcal{D} \cup \widetilde{\mathcal{D}})
  ~=~
  ( \pY(\mathcal{D}) \,\oplus\, \pY(\widetilde{\mathcal{D}})) \,/\, 2$.

\paragraph{Regression toward the mean.}
With the criterion \crit{same class}, it is obvious that
$\pY(\widetilde{\mathcal{D}})=\pY(\mathcal{D})$.
Therefore these variants of selective mixup are not concerned with resampling effects.%
\footnote{The absence of resampling effects holds for \crit{same class} and \crit{same domain} alone, but not in conjunction with other criteria. See e.g.\ the differences between \crit{\underline{same\phantom{y}domain}\,/\,diff. class} and \crit{\underline{any domain}\,/\,diff. class} in Figure~\ref{figWaterbirdsDistributions}.}
In contrast, the criteria \crit{different class} or \crit{different domain} do bias the sampling.
In the case of binary classification, 
we have
$\pY(\widetilde{\mathcal{D}}) \!=\! 1\!-\!\pY(\mathcal{D})$
and therefore
$\pY(\mathcal{D} \cup \widetilde{\mathcal{D}})$ is uniform.
This means that selective mixup with the \crit{different class} criterion has the side effect of balancing the training distribution of classes, a classical mitigation of 
class imbalance~\cite{japkowicz2000class,kubat1997addressing}.
For multiple classes, we have a more general result.

\begin{theorem}\label{thmRegression}
    Given a dataset
    $\mathcal{D} \!=\! \{(\bx_i,\by_i)\}_i$
    and paired data $\widetilde{\mathcal{D}}$
    sampled 
    according to the ``\crit{different class}'' criterion, i.e.\
    $\widetilde{\mathcal{D}} = \{ (\widetilde{\bx}_i, \widetilde{\by}_i) \sim \mathcal{D}
    ~~\textrm{s.t.}~~
    \widetilde{\by}_i\!\neq\!\by_i\}$,
    then the distribution of classes in
    $\mathcal{D} \mkern-0.6mu\cup\mkern-0.6mu \widetilde{\mathcal{D}}$
    is more uniform than in
    $\mathcal{D}$.
    Formally, the entropy\;\,
    $\entropyBig{\pY(\mathcal{D})} \,\le\, \entropyBig{\pY(\mathcal{D} \cup \widetilde{\mathcal{D}})}$.
    \\\textit{Proof:}~~\textup{see Appendix~\ref{appendixProof}.}\phantom{$\entropyBig{}$}
\end{theorem}
\vspace{-6pt}
Theorem~\ref{thmRegression} readily extends in two ways.
First, the same effect also results from the \crit{different domain} criterion: if each domain contains a different class distribution, the resampling from this criterion averages them out, yielding a more uniform aggregated training distribution.
Second, this averaging applies not only to class labels~($\by$) but also covariates~($\bx$).
An analysis using distributions is ill-suited but the mechanism similarly affects the sampling of covariates when training with selective mixup.

\paragraph{When does one benefit from the resampling (regardless of mixup)?}
The above results mean that selective mixup can implicitly reduce imbalances (a.k.a.\ biases) in the training data. 
When these are not spurious and also exist in the test data,
the effect on predictive performance could be detrimental.

We expect benefits (which we verify in Section~\ref{secExperiments})
on datasets with distribution shifts, whose training/test splits contain different imbalances by definition.
Softening imbalances in the training data is then likely to bring the training and test distributions closer to one another, in particular with extreme shifts such as the complete reversal of a spurious correlation (e.g.\ \dataset{waterbirds}~\cite{sagawa2019distributionally}, Section~\ref{expWatwerbirds}).

We also expect a benefit on worst-group metrics (e.g.\ with \dataset{civilComments}~\cite{koh2021wilds} in Section~\ref{expCivilComments}).
The challenge in these datasets comes from the imbalance of class/domain combinations in the training data, and prior work has indeed shown that balancing is a competitive baseline~\cite{idrissi2022simple,sagawa2019distributionally}.

\section{Experiments}
\label{secExperiments}

We performed a large number of experiments to understand the contribution
of the different effects of selective mixup and other resampling baselines (see Appendix~\ref{appendixResults} for complete results).

\paragraph{Datasets.}
We focus on five datasets that previously showed improvements with selective mixup.
We selected them to cover a range of modalities (vision, NLP, tabular),
settings (binary, multiclass),
and types of distribution shifts (covariate, label, and subpopulation shifts).
\begin{itemize}[topsep=-3pt,itemsep=3pt,labelindent=2pt,labelsep=*,leftmargin=12pt]
\item \textbf{Waterbirds}~\cite{sagawa2019distributionally} is a popular artificial dataset used to study distribution shifts. The task is to classify images of birds into two types. The image backgrounds are also of two types, and the correlation between birds and backgrounds is reversed across the training and test splits.
The type of background in each image serves as its domain label.

\item \textbf{CivilComments}~\cite{koh2021wilds} is a widely-used dataset of online text comments to be classified as toxic or not.
Each example is labeled with a topical attribute (e.g.\ Christian, male, LGBT, etc.) that is spuriously associated with ground truth labels in the training data.
These attributes serve as domain labels.
The target metric is the worst-group accuracy where the groups correspond to all toxicity/attribute combinations.

\item \textbf{Wild-Time Yearbook}~\cite{yao2022wild} contains yearbook portraits to be classified as male or female.
It is part of the Wild-Time benchmark, which is a collection of real-world datasets captured over time.
Each example belongs to a discrete time period that serves as its domain label.
Distinct time periods are assigned to the training and OOD test splits (see Figure~\ref{figRegression}).

\item \textbf{Wild-Time arXiv}~\cite{yao2022wild} contains titles of arXiv preprints. The task is to predict each paper's primary category among 172 classes. Time periods serve as domain labels.

\item \textbf{Wild-Time MIMIC-Readmission}~\cite{yao2022wild} contains hospital records (sequences of codes representing diagnoses and treatments) to be classified into two classes. The positive class indicates the readmission of the patient at the hospital within 15 days.
Time periods serve as domain labels.
\end{itemize}

\paragraph{Methods.}
We train standard architectures suited to each dataset with the methods below
(details in Appendix~\ref{appendixDetails}).
We perform early stopping i.e.\ recording metrics for each run at the epoch of highest ID or worst-group validation performance (for \dataset{Wild-Time} and \dataset{waterbirds}/\dataset{civilComments} datasets respectively).
We plot average metrics in bar charts over 9 different seeds with error bars representing $\pm$ one standard deviation.
\textbf{ERM} and \textbf{vanilla mixup} are the standard baselines.
Baseline \textbf{resampling}
uses training examples with equal probability from each class, domain, or combinations thereof
as in~\cite{idrissi2022simple,sagawa2019distributionally}.
\textbf{Selective mixup}~(\textcolor{myRed}{$\blacksquare$})
includes all possible selection criteria based on classes and domains.
We avoid ambiguous terminology from earlier works because of inconsistent usage (e.g.\ ``intra-label LISA'' means ``different domain'' in~\cite{koh2021wilds} but not in~\cite{yao2022wild}).
\textbf{Selective sampling}~(\textcolor{myBlue}{$\blacksquare$})
is a novel ablation of selective mixup where the selected pairs are not mixed, but the instances are appended one after another in the mini-batch. Half are dropped at random to keep the mini-batch size identical to the other methods.
Therefore any difference between selective sampling and ERM is attributable only to resampling effects.
We also include \textbf{novel combinations}~(\textcolor{myMagenta}{$\blacksquare$}) of sampling and mixup.

\subsection{Results on the \dataset{waterbirds} dataset}
\label{expWatwerbirds}

\begin{figure*}[h!]
  \includegraphics[width=6.4cm]{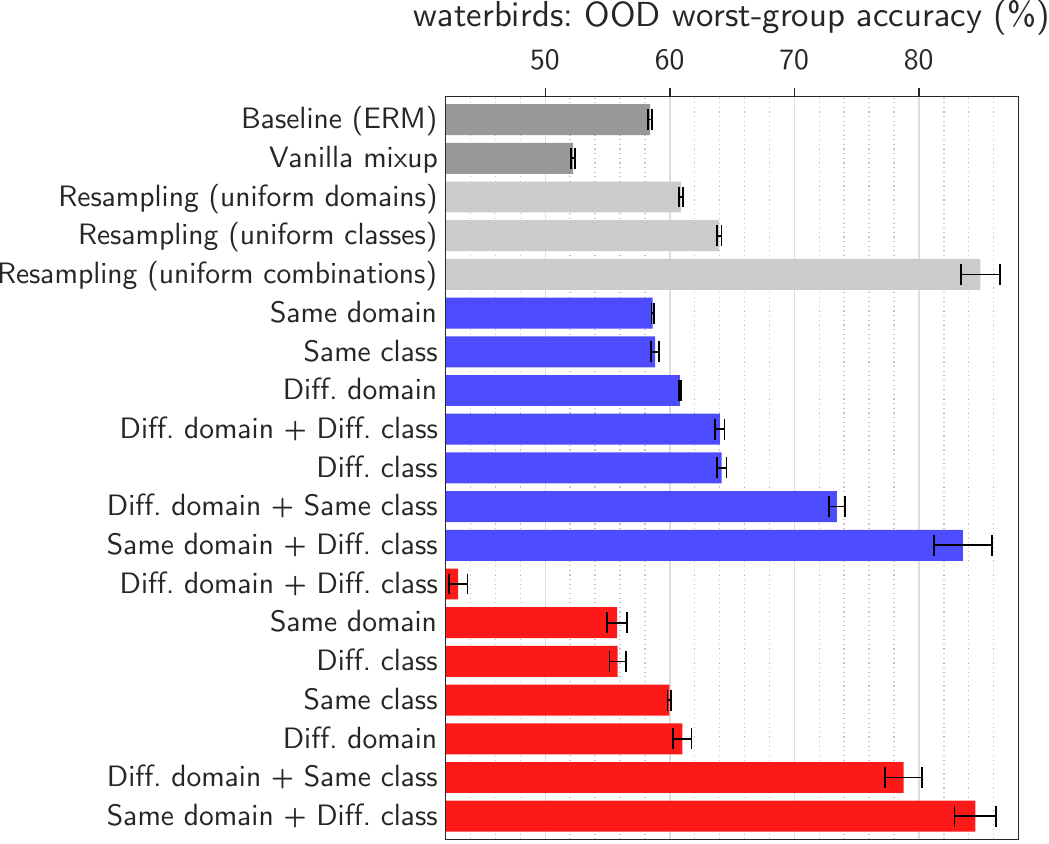}
  \caption{Main results on \dataset{waterbirds}.\vspace{6pt}\\
  We first observe that vanilla mixup is detrimental compared to ERM.
  Resampling with uniform class/domain combinations is hugely beneficial, for the reasons explained in Figure~\ref{figWaterbirdsDistributions}.
  The ranking of various criteria for selective sampling is similar whether with or without mixup.
  Most interestingly, the best criterion performs similarly, but no better than the best resampling.
  \vspace{6pt}\\
  \fbox{\begin{minipage}{10.85em}
  \textcolor{myGray}{$\blacksquare$}~\textcolor{myDarkGray}{\scriptsize{Baselines}}\vspace{-2pt}\\
  \textcolor{myBlue}{$\blacksquare$~\scriptsize{Selective sampling without mixup}}\vspace{-2pt}\\
  \textcolor{myRed}{$\blacksquare$~\scriptsize{Selective mixup}}\end{minipage}}
  }
  \label{figWaterbirds}
\end{figure*}

The target metric for this dataset is the worst-group accuracy, with groups defined as the four class/domain combinations.
The two difficulties are (1)~a class imbalance (77\,/\,23\%) and (2)~a correlation shift (spurious class/domain association reversed at test time).
See discussion in Figure~\ref{figWaterbirds}.
\vspace{-4pt}

\floatsetup[figure]{margins=hangright,capposition=beside,capbesideposition=right,floatwidth=0.5\textwidth,capbesidewidth=0.48\textwidth}

\begin{figure*}[h!]
  \includegraphics[width=.99\linewidth]{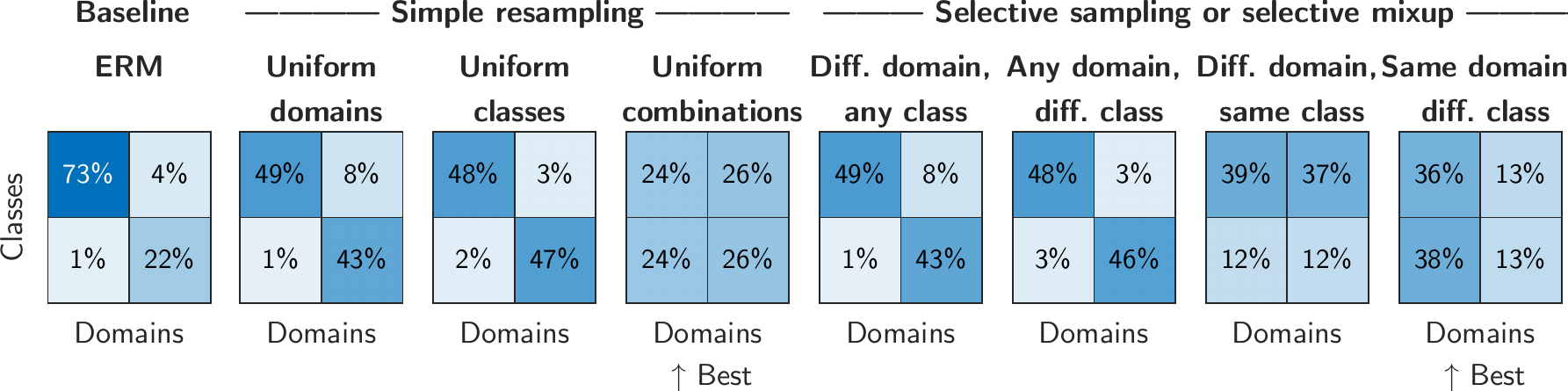}
  \scriptsize{\textbf{Resampling uniform combinations} gives them all equal weights, just like the worst-group target metric.\\
  \textbf{Selective mixup with same domain\,/\,diff. class} also gives equal weights to the classes, while breaking \\the spurious pattern between groups and classes, unlike any other criterion.}
  \caption{The sampling ratios of each class/domain clearly explain the performance of the best methods (\dataset{waterbirds}).\label{figWaterbirdsDistributions}}
\end{figure*}

\vspace{-4pt}
This suggest that \textbf{the excellent performance of the best version of selective mixup is here entirely due to resampling}.
Note that the efficacy of resampling on this dataset is not a new finding~\cite{idrissi2022simple,sagawa2019distributionally}.
What is new is its equivalence with the best variant of selective mixup.
We further explain this claim in Figure~\ref{figWaterbirdsDistributions}
by examining the proportions of classes and domains sampled by each training method.

\floatsetup[figure]{margins=hangright,capposition=beside,capbesideposition=right,floatwidth=0.75\textwidth,capbesidewidth=0.22\textwidth}

\subsection{Results on the \dataset{yearbook} dataset}

\begin{figure*}[h!]
  \includegraphics[width=7.2cm]{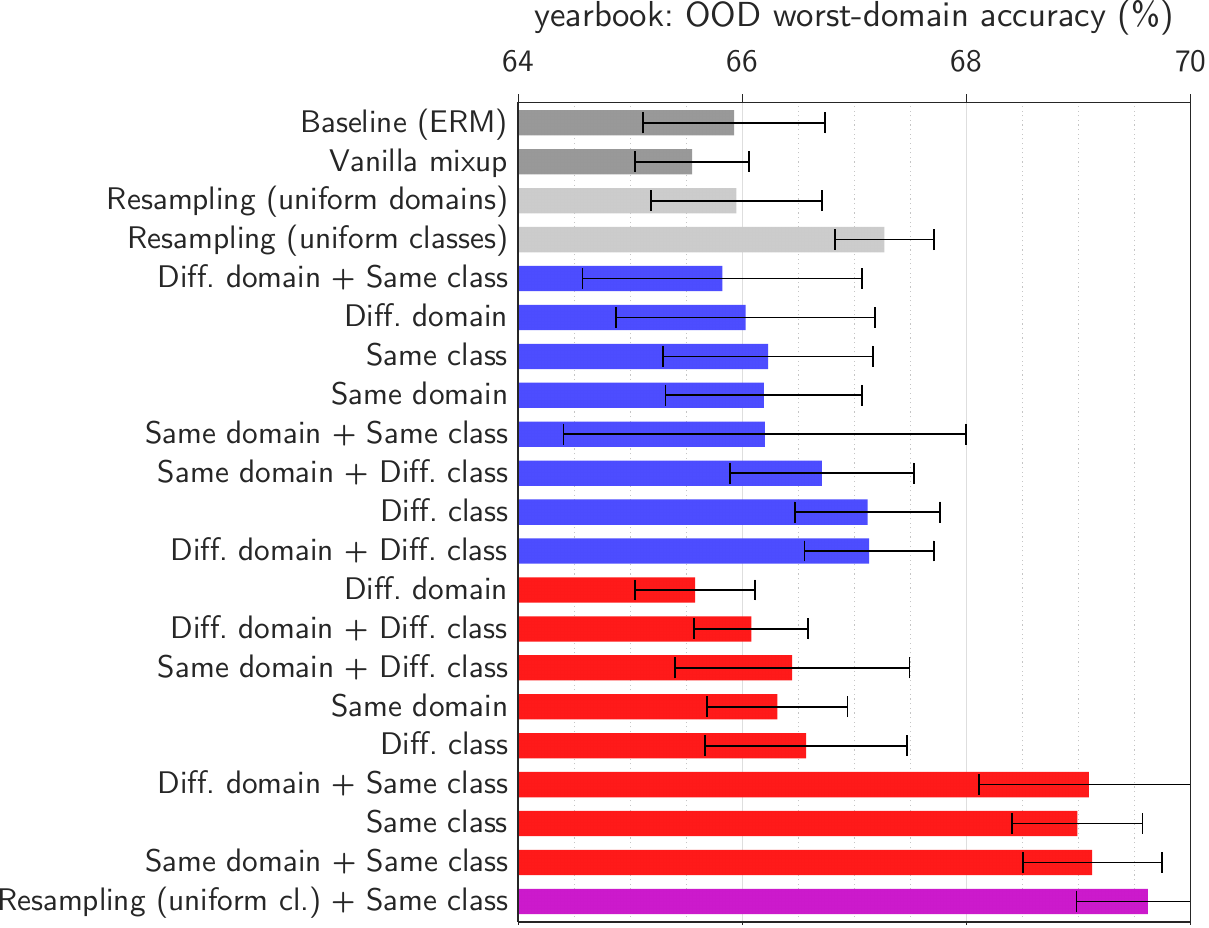}
  \caption{Main results on \dataset{yearbook}.\vspace{10pt}\\
  With selective mixup, the ``different class'' criterion is not useful, but ``same class'' performs significantly better than ERM.
  Since this criterion alone does not have resampling effects, 
  it indicates a genuine benefit from mixup restricted to pairs of the same class.
  \vspace{10pt}\\
  \fbox{\begin{minipage}{13.7em}
  \textcolor{myGray}{$\blacksquare$}~\textcolor{myDarkGray}{\scriptsize{Baselines}}\vspace{-2pt}\\
  \textcolor{myBlue}{$\blacksquare$~\scriptsize{Selective sampling without mixup}}\vspace{-2pt}\\
  \textcolor{myRed}{$\blacksquare$~\scriptsize{Selective mixup}}\vspace{-2pt}\\
  \textcolor{myMagenta}{$\blacksquare$~\scriptsize{Novel combinations of sampling and mixup}}\end{minipage}}
  }
  \label{figYearbook}
\end{figure*}

The difficulty of this dataset comes from a slight class imbalance and the presence of covariate/label shift (see Figure~\ref{figRegression}).
The test split contains several domains (time periods). The target metric is the worst-domain accuracy.
Figure~\ref{figYearbook} shows that vanilla mixup is slightly detrimental compared to ERM.
Resampling for uniform classes gives a clear improvement because of the class imbalance.
With selective sampling (no mixup), the only criteria that improve over ERM contain ``different class''.
This is expected because this criterion implicitly resamples for a uniform class distribution.

\floatsetup[figure]{margins=hangright,capposition=beside,capbesideposition=right,floatwidth=0.53\textwidth,capbesidewidth=0.44\textwidth}

To investigate whether some of the improvements are due to resampling,
we measure the divergence between training and test distributions of classes and covariates (details in Appendix~\ref{appendixDetails}).
Figure~\ref{figYearbook2}) shows first that there is a clear variation among different criteria (\textcolor{myBlue}{$\bullet$ blue dots}) i.e.\ some bring the training/test distributions closer to one another. Second, there is a remarkable correlation between the test accuracy and the divergence, on both classes and covariates.%
\footnote{\label{noteDomainsYearbook}As expected, the correlation is reversed for the first two test domains in Figure~\ref{figYearbook2}
since they are even further from a uniform class distribution than the average of the training data,
as seen in Figure~\ref{figRegression}.}
This means that resampling effects do occur and also play a part in the best variants of selective mixup.

\floatsetup[figure]{margins=centering,capposition=bottom,floatwidth=\textwidth}
\begin{figure*}[h!]
  \includegraphics[width=.98\textwidth]{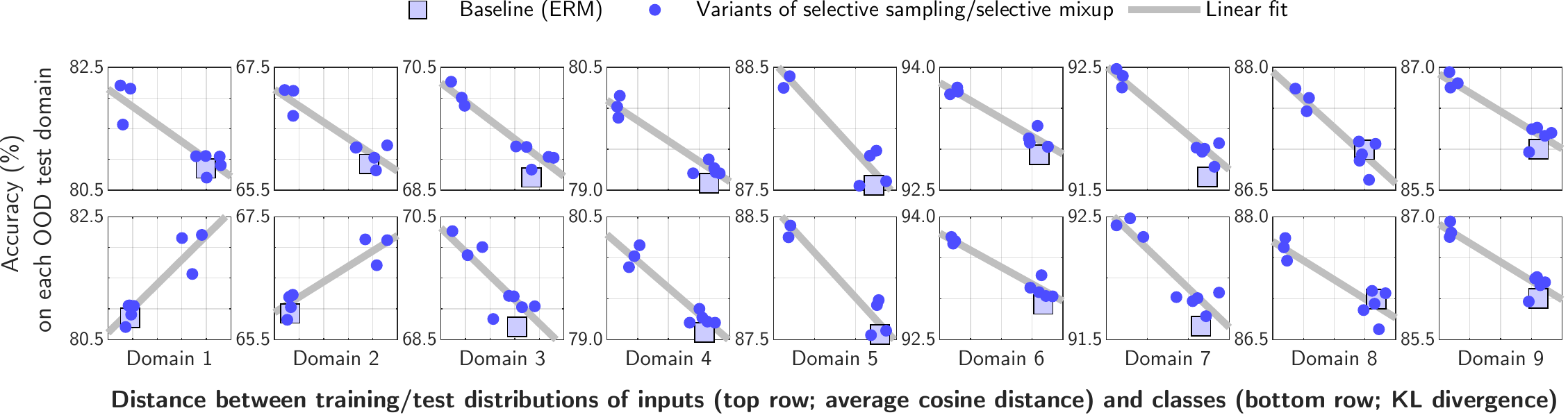}
  \caption[abc]{
  Different selection criteria (\textcolor{myBlue}{$\bullet$}) modify the distribution of both covariates and labels (upper and lower rows).
  The resulting reductions in divergence between training and test distributions correlate remarkably well with test performance.\footnotemark[3]
  This confirms the contribution of resampling to the overall performance of selective mixup.
  \label{figYearbook2}}
\end{figure*}

Finally, the improvements from simple resampling and the best variant of selective mixup suggest a new combination.
We train a model with uniform class sampling and selective mixup using the ``same class'' criterion, and obtain performance superior to all existing results (last row in Figure~\ref{figYearbook2}).
This confirms the \textbf{complementarity of the effects of resampling and within-class selective mixup}.

\subsection{Results on the \dataset{arXiv} dataset}

\begin{figure*}[h!]
  \includegraphics[width=8.0cm]{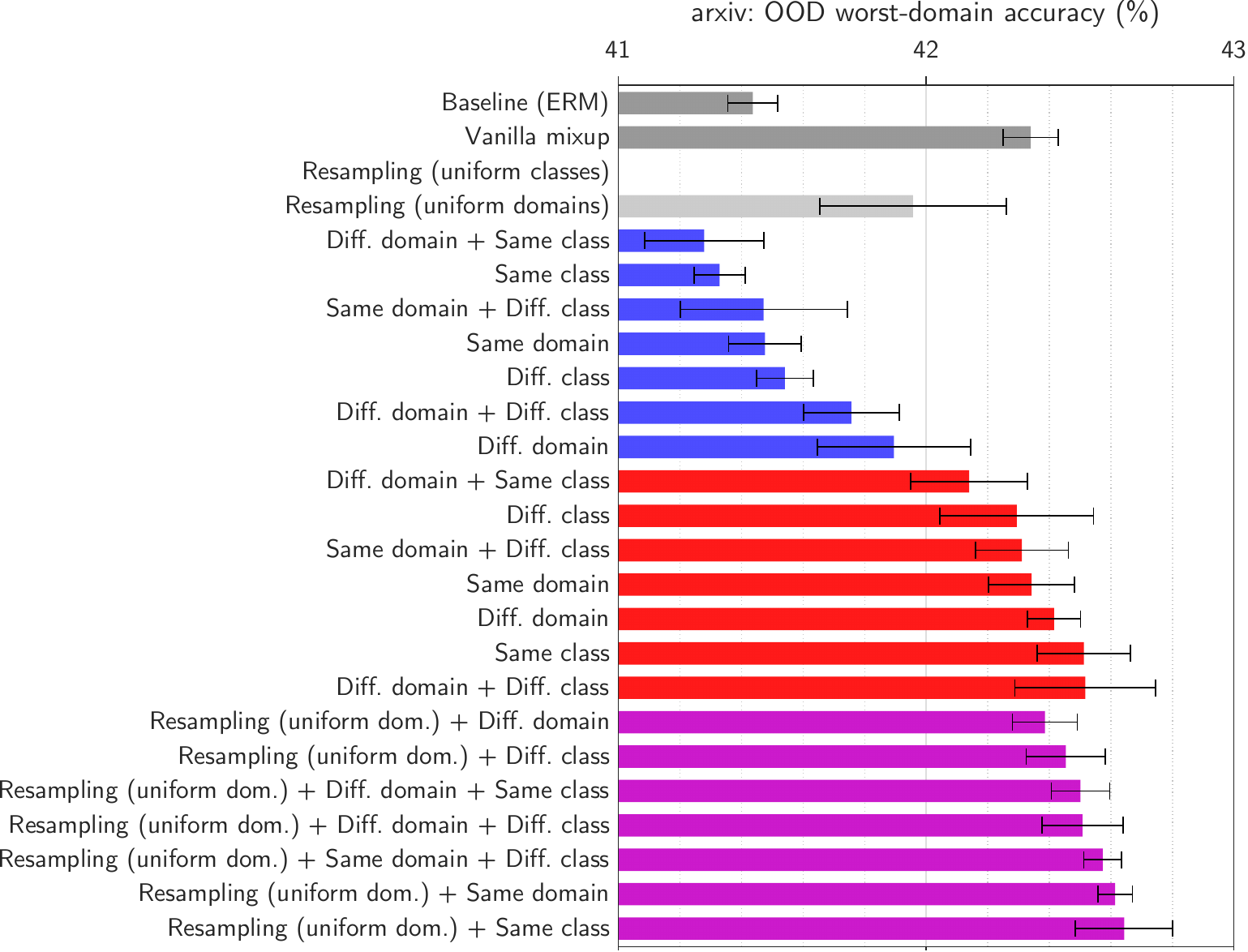}
  \caption{Main results on \dataset{arXiv}.\vspace{4pt}\\
  \fbox{\begin{minipage}{13.7em}
  \textcolor{myGray}{$\blacksquare$}~\textcolor{myDarkGray}{\scriptsize{Baselines}}\vspace{-2pt}\\
  \textcolor{myBlue}{$\blacksquare$~\scriptsize{Selective sampling without mixup}}\vspace{-2pt}\\
  \textcolor{myRed}{$\blacksquare$~\scriptsize{Selective mixup}}\vspace{-2pt}\\
  \textcolor{myMagenta}{$\blacksquare$~\scriptsize{Novel combinations of sampling and mixup}}\end{minipage}}
  \vspace{6pt}\\
  To investigate the contribution of resampling, we measure the divergence between training/test class distributions and plot them against the test accuracy
  (Figure~\ref{figArxiv2}).
  We observe a strong correlation across methods.
  Mixup essentially offsets the performance by a constant factor. 
  This suggests again the independence of the effects of mixup and resampling.
  \label{figArxiv}}
\end{figure*}

This dataset has difficulties similar to \dataset{yearbook} and also many more classes (172).
Simple resampling for uniform classes is very bad (literally off the chart in Figure~\ref{figArxiv}) because it overcorrects the imbalance (the test distribution being closer to the training than to a uniform one).
Uniform \emph{domains} is much better since its effect is similar but milder.

All variants of selective mixup (\textcolor{myRed}{$\blacksquare$})
perform very well, but they improve over ERM even without mixup (\textcolor{myBlue}{$\blacksquare$}).
And the selection criteria rank similarly with or without mixup, suggesting that parts of the improvements of selective mixup is due to the resampling.
Given that vanilla mixup also clearly improves over ERM, the performance of \textbf{selective mixup is explained by cumulative effects of vanilla mixup and resampling effects}.
This also suggests new combinations of methods (\textcolor{myMagenta}{$\blacksquare$}) 
among which we find one version marginally better than the best variant of selective mixup (last row).

\floatsetup[figure]{margins=hangright,capposition=beside,capbesideposition=right,floatwidth=0.58\textwidth,capbesidewidth=0.39\textwidth}

\floatsetup[figure]{margins=hangright,capposition=beside,capbesideposition=right,floatwidth=0.31\textwidth,capbesidewidth=0.66\textwidth}
\begin{figure*}[h!]
  \includegraphics[height=6.1cm]{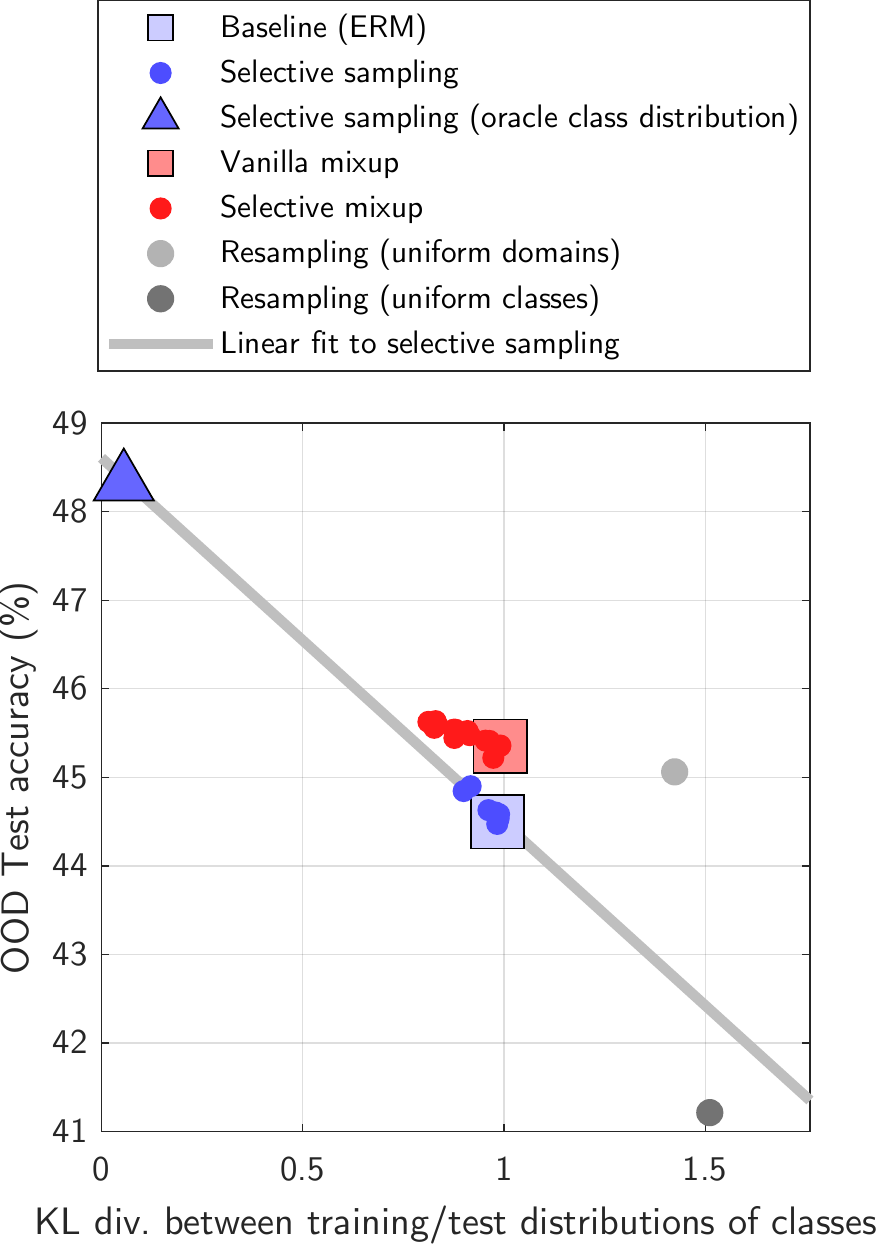}
  \caption{Divergence of tr./test class distributions vs. test accuracy.
  \vspace{6pt}\\
  The resampling baselines (\textcolor{myGray}{$\bullet$}\textcolor{myDarkGray}{$\bullet$}) also roughy agree with a linear fit to the ``selective sampling'' points.
  We therefore hypothesize that \textbf{all these methods are mostly addressing label shift}.
  We verify this hypothesis with the remarkable fit of an additional point (\textcolor{myLightBlue}{$\blacktriangle$})
  of a model trained by resampling according to the \underline{test} set class distribution, i.e.\ cheating.
  \vspace{6pt}\\
  It represents an upper bound that might be achievable in future work
  with methods for label shift~\cite{azizzadenesheli2019regularized,lipton2018detecting}.
  \vspace{6pt}\\
  We replicated these observations on every test domain of this dataset (Figure~\ref{figAppendixArxiv2} in the appendix).
  \label{figArxiv2}}
\end{figure*}

\subsection{Results on the \dataset{civilComments} dataset}
\vspace{-3pt}
\label{expCivilComments}

This dataset mimics a subpopulation shift because the worst-group metric
requires high accuracy on classes and domains under-represented in the training data.
It also contains an implicit correlation shift because any class/domain association (e.g.\ ``Christian'' comments labeled as toxic more often than not) becomes spurious when evaluating individual class/domain combinations.

\floatsetup[figure]{margins=hangright,capposition=beside,capbesideposition=right,floatwidth=0.56\textwidth,capbesidewidth=0.42\textwidth}
\begin{figure*}[h!]
  \includegraphics[width=7.8cm]{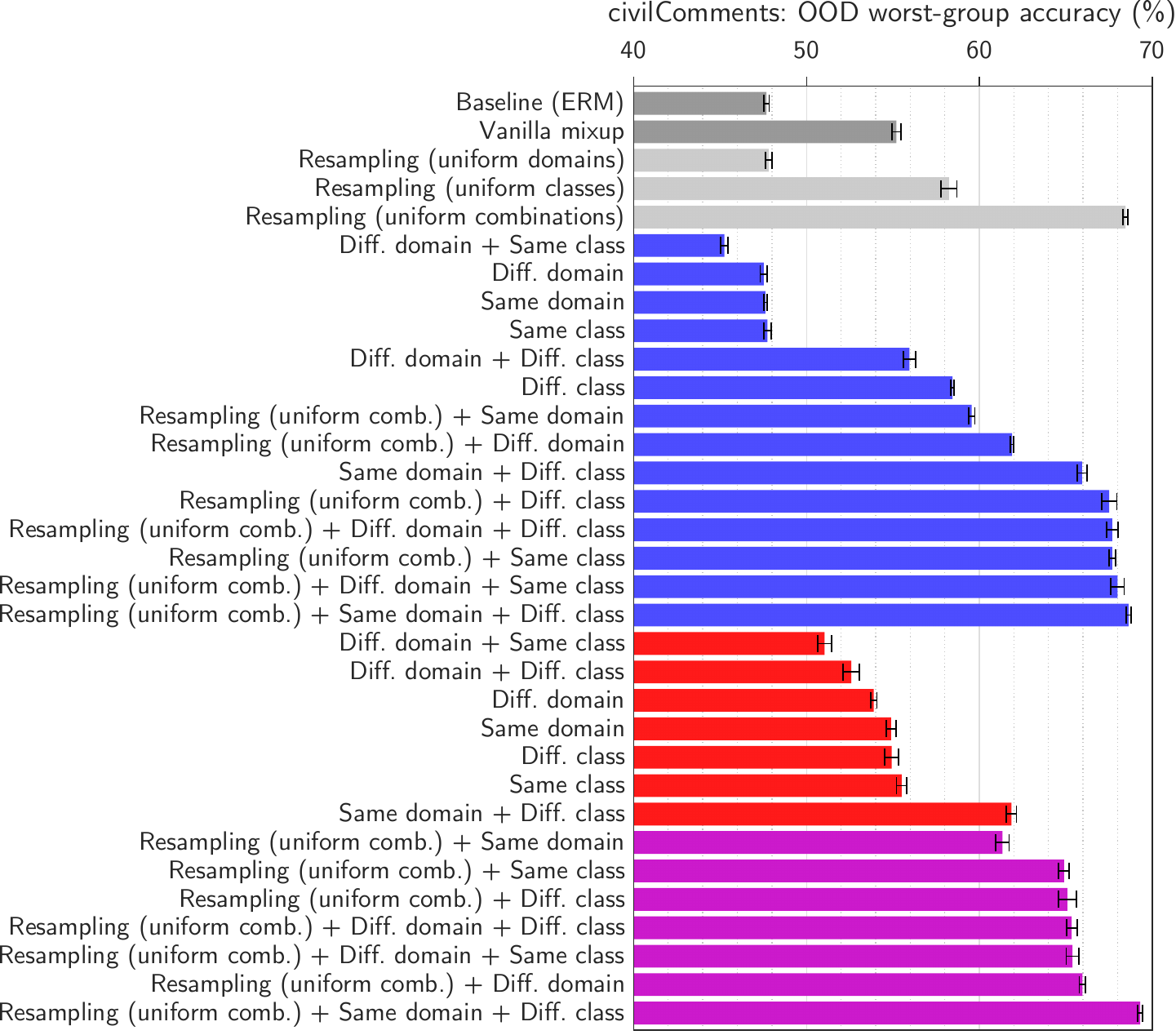}
  \caption{Main results on \dataset{civilComments}.\vspace{6pt}\\
  For the above reasons, it makes sense that resampling for uniform classes or combinations greatly improves performance,
  as shown in prior work~\cite{idrissi2022simple}.
  \vspace{6pt}\\
  With selective mixup (\textcolor{myRed}{$\blacksquare$}), some criterion (same domain/diff. class) performs clearly above all others.
  But it works \textbf{even better without mixup}! (\textcolor{myBlue}{$\blacksquare$})
  Among many other variations, \textbf{none surpasses the uniform-combinations baseline}.
  \vspace{11pt}\\
  \fbox{\begin{minipage}{13.7em}
  \textcolor{myGray}{$\blacksquare$}~\textcolor{myDarkGray}{\scriptsize{Baselines}}\vspace{-2pt}\\
  \textcolor{myBlue}{$\blacksquare$~\scriptsize{Selective sampling without mixup}}\vspace{-2pt}\\
  \textcolor{myRed}{$\blacksquare$~\scriptsize{Selective mixup}}\vspace{-2pt}\\
  \textcolor{myMagenta}{$\blacksquare$~\scriptsize{Novel combinations of sampling and mixup}}\end{minipage}}
  }
  \label{figCivilComments}
\end{figure*}

\subsection{Results on the \dataset{MIMIC-Readmission} dataset}

This dataset contains a class imbalance (about 78/22\% in training data), label shift (the distribution being more  balanced in the test split), and possibly covariate shift. It is unclear whether the task is causal or anticausal (labels causing the features) because the inputs contain both diagnoses and treatments. 
The target metric is the area under the ROC curve (AUROC)
which gives equal importance to both classes.
We report the worst-domain AUROC, i.e.\ the lowest value across test time periods.

\begin{figure*}[h!]
  \includegraphics[width=6.5cm]{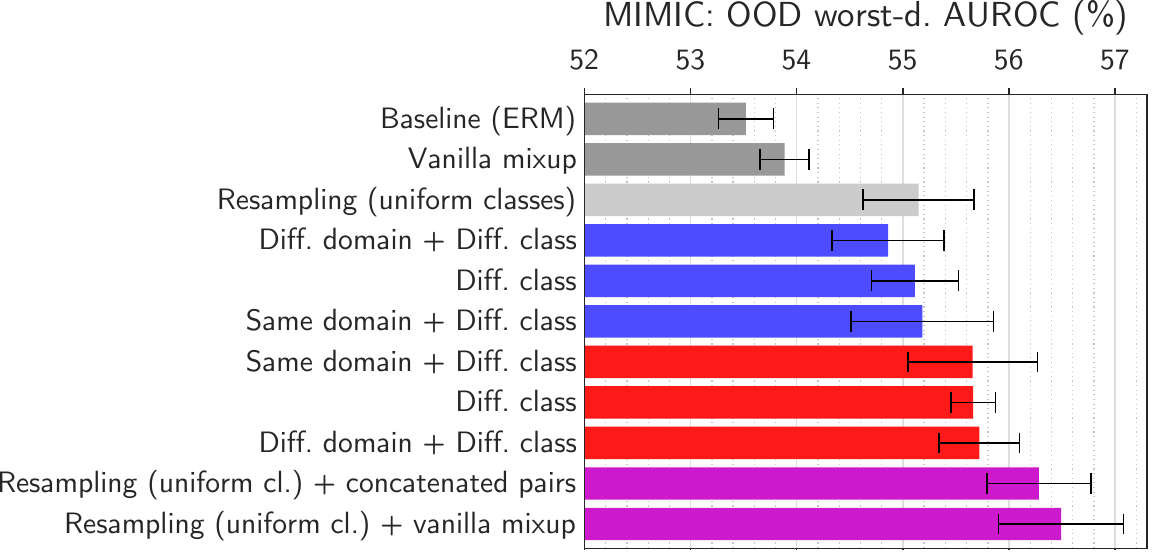}
  \caption{Main results on \dataset{MIMIC-Readmission}.\vspace{6pt}\\
  \fbox{\begin{minipage}{13.7em}
  \textcolor{myGray}{$\blacksquare$}~\textcolor{myDarkGray}{\scriptsize{Baselines}}\vspace{-2pt}\\
  \textcolor{myBlue}{$\blacksquare$~\scriptsize{Selective sampling without mixup}}\vspace{-2pt}\\
  \textcolor{myRed}{$\blacksquare$~\scriptsize{Selective mixup}}\vspace{-2pt}\\
  \textcolor{myMagenta}{$\blacksquare$~\scriptsize{Novel combinations of sampling and mixup}}\end{minipage}}
  }
  \label{figMimic}
\end{figure*}

Vanilla mixup performs a bit better than ERM.
Because of the class imbalance, resampling for uniform classes also improves ERM.
As expected, this is perfectly equivalent to the selective sampling criterion ``diffClass'' and they perform therefore equally well.
Adding mixup is yet a bit better, which suggests again that \textbf{the performance of selective mixup
is merely the result of the independent effects of vanilla mixup and resampling}.
We further verify this explanation with the novel combination of simple resampling and vanilla mixup, and observe almost no difference whether the mixing operation is performed or not (last two rows in Figure~\ref{figMimic}).

\floatsetup[figure]{margins=hangright,capposition=beside,capbesideposition=right,floatwidth=0.5\textwidth,capbesidewidth=0.48\textwidth}

To further support the claim that these methods mostly address label shift, we report in Table~\ref{tabMimic}
the proportion of the majority class in the training and test data. We observe that the distribution sampled by the best training methods brings it much closer to that of the test data.

\floatsetup[table]{margins=hangright,capposition=beside,capbesideposition=right,floatwidth=0.44\textwidth,capbesidewidth=0.53\textwidth}
\begin{table}[h!]
    \caption{The performance of the various methods on \dataset{MIMIC-Readmission} is
    explained by their correction of a class imbalance.
    The best training methods (boxed numbers) sample the majority class in a proportion much closer to that of the test data.\label{tabMimic}\vspace{20pt}}
    \centering
    \renewcommand{\tabcolsep}{0.5em}
    \renewcommand{\arraystretch}{0.95}
    \scriptsize
    \begin{tabular}{lc}
    \toprule
    \textbf{Proportion of majority class} & \textbf{(\%)}\\
    \midrule
    In the dataset (training) & 78.2\\
    In the dataset (validation) & 77.8\\
    In the dataset (OOD test) & \boxStart{1}66.5\boxEnd{1}\\
    \midrule
    \textbf{Sampled by different training methods}\\
      Resampling (uniform classes) & 50.0\\
      Diff. domain + diff. class & 50.0\\
      Diff. class & 50.1\\
      Same domain + Diff. class & 49.9\\
      Resampling (uniform cl.) + concatenated pairs & \boxStart{2}64.3\\
      Resampling (uniform cl.) + vanilla mixup & 64.3\boxEnd{2}\\
    \bottomrule
    \end{tabular}
\end{table}

\subsection{Evidence of a ``regression toward the mean'' in the data}
\label{secExperimentsRegression}

\begin{figure*}[h!]
  \includegraphics[width=.32\linewidth]{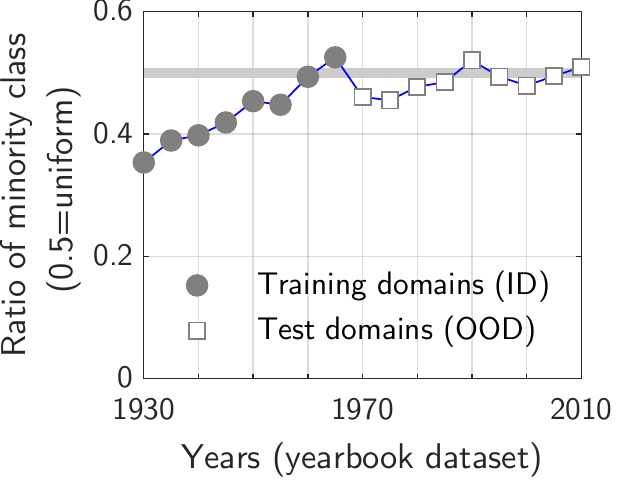}~~
  \includegraphics[width=.32\linewidth]{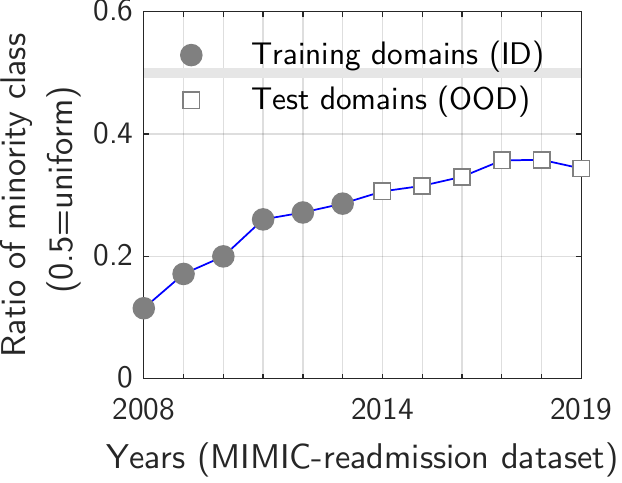}~~
  \includegraphics[width=.32\linewidth]{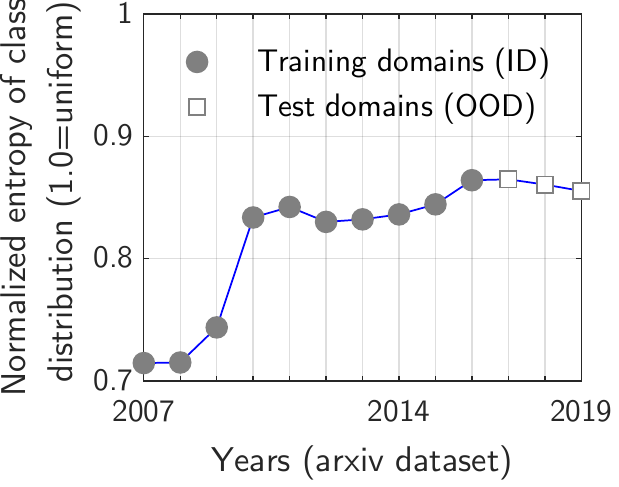}
  \caption{The class distribution shifts toward uniformity in these Wild-Time datasets.
  This agrees with the explanation that the resampling benefits rely on a ``regression toward the mean''.}
  \label{figRegression}
\end{figure*}

We hypothesized in Section~\ref{secBias} that the resampling benefits are due to a ``regression toward the mean'' across training and test splits.
We now check for this property
and find indeed a shift toward uniform class distributions in all datasets studied.
For the Wild-Time datasets, we plot in Figure~\ref{figRegression} the ratio of the minority class (binary tasks: yearbook, MIMIC) and class distribution entropy (multiclass task: arxiv).
Finding this property in all three datasets agrees with the proposed explanation
and the fact that we selected them because they previously showed improvements with selective mixup in~\cite{yao2022wild}.

In the \dataset{waterbirds} and \dataset{civilComments} datasets, the shift toward uniformity¨also holds, but artificially.
The training data contains imbalanced groups (class/domain combinations) whereas the evaluation with worst-group accuracy implicitly gives uniform importance to all groups.

\floatsetup[figure]{margins=hangright,capposition=beside,capbesideposition=right,floatwidth=0.73\textwidth,capbesidewidth=0.25\textwidth}

\section{Related work}

\paragraph{Mixup and variants.}
Mixup was originally introduced in~\cite{mixup} and
numerous variants followed~\cite{cao2022survey}.
Many propose modality-specific mixing operations:
CutMix~\cite{cutmix} replaces linear combinations with collages of image patches,
Fmix~\cite{fmix} combines image regions based on frequency contents, 
AlignMixup~\cite{venkataramanan2022alignmixup} combines images after spatial alignment. 
Manifold-mixup~\cite{manifold_mix} replaces the mixing in input space with the mixing of learned representations,
making it applicable to
text embeddings.

\paragraph{Mixup for OOD generalization.}
Mixup has been integrated into existing techniques for 
domain adaptation (DomainMix~\cite{xu2020adversarial}),
domain generalization (FIXED~\cite{lu2022fixed}),
and with meta learning (RegMixup~\cite{pinto2022regmixup}).
This paper focuses on variants we call ``selective mixup'' that use non-uniform sampling of the pairs of mixed examples.
LISA~\cite{yao2022improving} proposes two heuristics, same-class/different-domain and vice versa, used in proportions tuned by cross-validation on each dataset.
\citet{palakkadavath2022improving} use same-class pairs and an additional objective to encourage invariance of the representations to the mixing.
CIFair~\cite{tian2023cifair} uses same-class pairs with a contrastive objective to improve algorithmic fairness.
SelecMix~\cite{hwang2022selecmix} proposes a selection heuristic to handle biased training data: same class/different biased attribute, or vice versa.
DomainMix~\cite{xu2020adversarial} uses different-domain pairs for domain adaptation.
DRE~\cite{li2023data} uses same-class/different-domain pairs and regularize their Grad-CAM explanations to improve OOD generalization.
SDMix~\cite{lu2022semantic} applies mixup on examples from different domains with other improvements to improve cross-domain generalization for activity recognition.

\paragraph{Explaining the benefits of mixup}
has invoked regularization~\cite{zhang2020does} and augmentation~\cite{kimura2021mixup} effects, the introduction of label noise~\cite{liu2023over}, and the learning of rare features~\cite{zou2023benefits}.
These works focus on the mixing and in-domain generalization, whereas we focus on the selection and OOD generalization.

\paragraph{Training on resampled data.}
We find that selective mixup is sometimes equivalent to training on resampled or reweighted data.
Both are standard tools~\cite{japkowicz2000class,kubat1997addressing} to handle distribution shifts in a domain adaptation setting, also known as importance-weighted empirical risk minimization (IW-ERM)~\cite{shimodaira2000improving,gretton2009covariate}.
For covariate shift,
IW-ERM trains a model with a weight or sampling probability on each training point $\bx$ as its likelihood ratio $p_\textrm{target}(\bx)/p_\textrm{source}(\bx)$.
Likewise with labels $\by$ and $p_\textrm{target}(\by)/p_\textrm{source}(\by)$
for label shift~\cite{azizzadenesheli2019regularized,lipton2018detecting},
Recently, \cite{idrissi2022simple,sagawa2019distributionally} showed that reweighting and resampling are competitive with the state of the art on multiple OOD and label-shift benchmarks~\cite{garg2023rlsbench}.

\section{Conclusions and open questions}

\paragraph{Conclusions.}
This paper helps understand selective mixup, which is one of the most successful and general methods for distribution shifts.
We showed unambiguously that much of the improvements were actually unrelated to the mixing operation and could be obtained with much simpler, well-known resampling methods.
On datasets where mixup does bring benefits, we could then
obtain even better results by combining the independent effects of the best mixup and resampling variants.

\paragraph{Limitations.}
We focused on the simplest version selective mixup as described by \citet{yao2022improving}.
Many papers combine the principle with modifications to the learning objective~\cite{hwang2022selecmix,li2023data,lu2022semantic,palakkadavath2022improving,tian2023cifair,xu2020adversarial}.
Resampling likely plays a role in these methods too but this claim requires further investigation.

We evaluated ``only'' five datasets. Since we introduced simple ablations that can single out the effects of resampling, we hope to see future re-evaluations of other datasets.

Because we picked datasets that had previously shown benefits with selective mixup,
we could not fully verify the predicted failure when there is no ``regression toward the mean'' in the data.
We present one experiment verifying this prediction on \dataset{yearbook} by swapping the ID and OOD data (see Appendix~\ref{appendixPredictedFailure}).

Finally, this work is \textbf{not} about designing new algorithms to surpass the state of the art.
Our focus is on improving the scientific understanding of existing mixup strategies and their limitations.

\paragraph{Open questions.}
Our results leave open the question of the applicability of selective mixup to real situations.
The ``{regression toward the mean}'' explanation indicates that much of the observed improvements are accidental since they rely on an artefact of some datasets.
In real deployments, distribution shifts cannot be foreseen in nature nor magnitude.
This is a reminder of the relevance of Goodhart's law to machine learning~\cite{teney2020value} and of the risk of overfitting to popular benchmarks~\cite{liao2021we}.


\clearpage
\bibliography{main}
\bibliographystyle{plainnat} 

\clearpage

\appendix
\section*{\Large Appendices}

\section{Experimental details}
\label{appendixDetails}

We follow prior work on each dataset for the \textbf{architectures and hyperparameters} of our experiments. 
For each dataset, all methods compared use hyperparameters initially validated with the ERM baseline.
All experiments use early stopping i.e.\ recording metrics for each run at the epoch of highest ID or worst-group validation performance (for \dataset{Wild-Time} and \dataset{waterbirds}/\dataset{civilComments} datasets respectively).
Each dataset/method is run with 9 different seeds unless otherwise noted.
The bar charts report the average over these seeds and error bars represent $\pm$ one standard deviation.

We noticed considerable \textbf{variability in the results reported in prior work}, sometimes for datasets/methods supposedly identical (e.g.\ resampling baselines on \dataset{waterbirds}). Therefore we only 
make comparisons across results obtained within a unique code base after re-running all baselines in the same setting.

We also found some \textbf{issues in existing code} that we could not clear up with their authors despite multiple requests. This includes inconsistent preprocessing and duplicated data in the preprocessing of civilComments in~\cite{idrissi2022simple}, ``magic constants'' in the implementation of selective mixup (LISA) in~\cite{yao2022improving}, inappropriate architectures for \dataset{MIMIC} in \cite{yao2022wild}.
We fixed these issues in our codebase.
Therefore we refrain from claims or direct comparisons with the absolute state of the art.

\begin{figure*}[h!]
  \includegraphics[width=.8\textwidth]{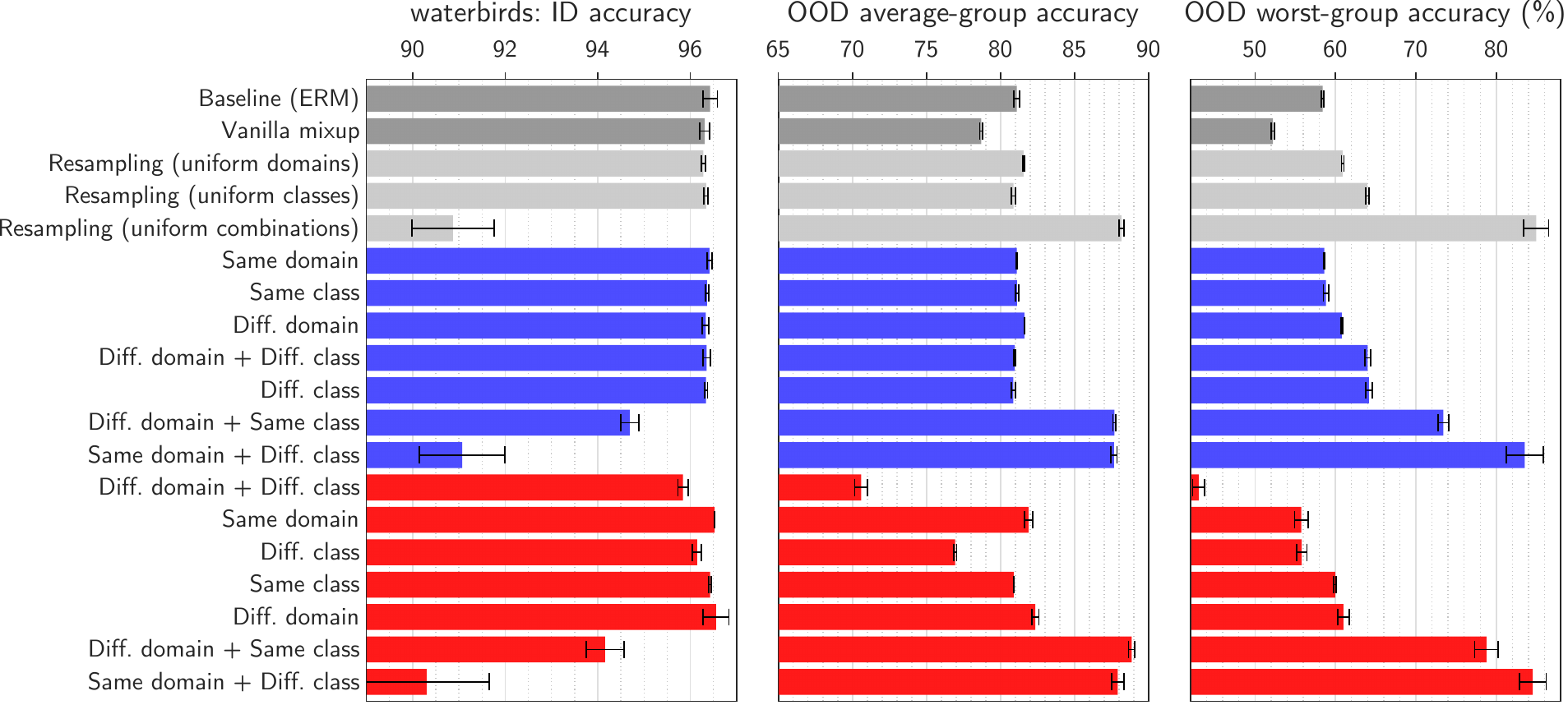}\vspace{14pt}\\
  \includegraphics[width=.815\textwidth]{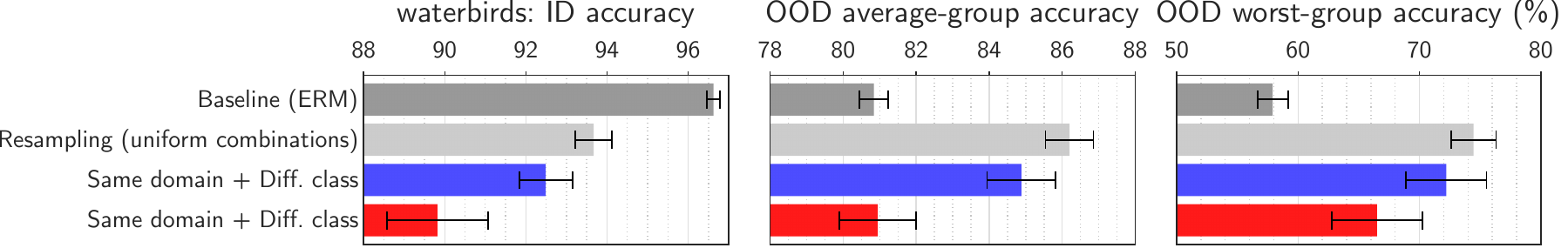}
  \caption{Results on \dataset{waterbirds} (top) with linear classifiers on frozen ResNet-50 features and
  (bottom) with fine-tuned ResNet-50 models (selected methods only).
  \label{figAppendixWaterbirds}}
\end{figure*}

\begin{figure*}[h!]
  \includegraphics[width=.75\textwidth]{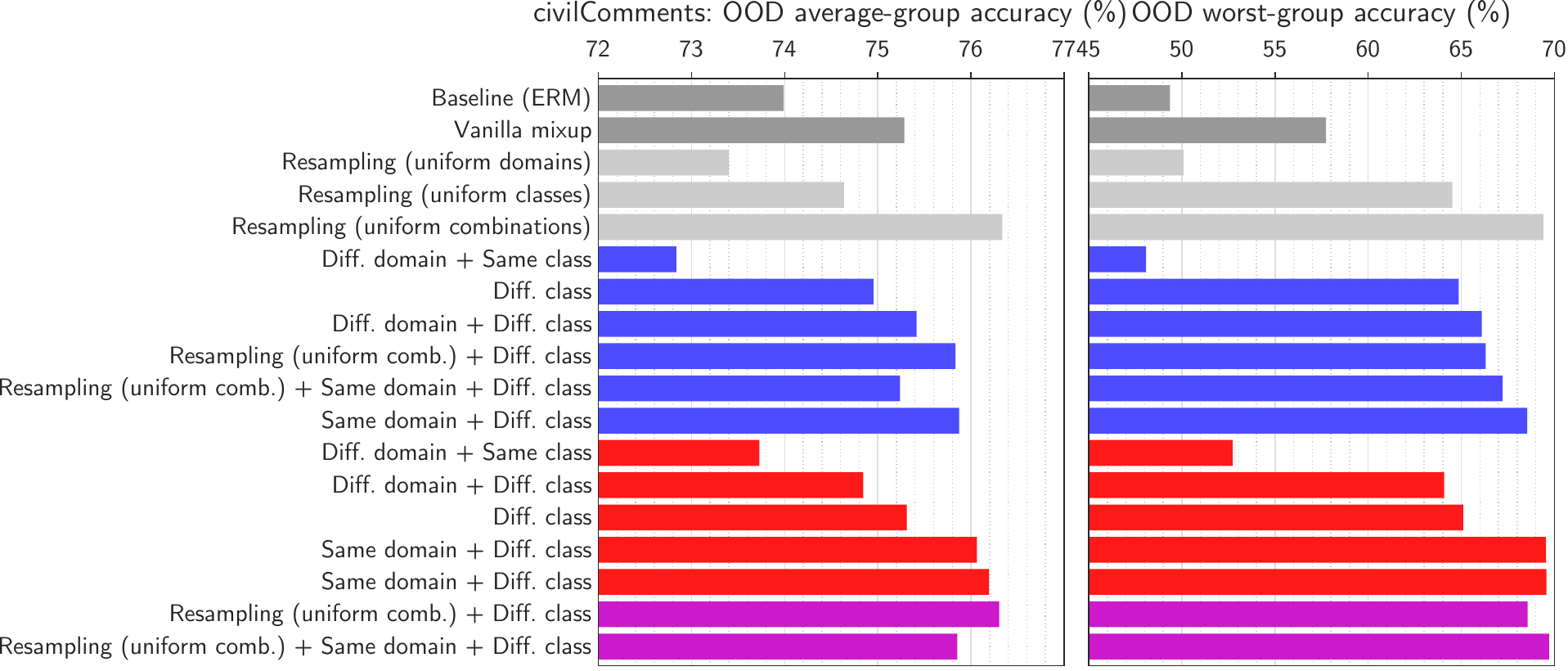}
  \caption{Results on \dataset{civilComments} with fine-tuned BERT models (single seed, reduced set of methods).
  These results are qualitatively identical to those with frozen embeddings above.
  \label{figAppendixCivilComments2}}
\end{figure*}

Dataset-specific notes:
\begin{itemize}[topsep=-3pt,itemsep=3pt,labelindent=2pt,labelsep=*,leftmargin=12pt]
\item On \dataset{waterbirds}, we use ImageNet-pretrained ResNet-50 models. The results in the main paper use linear classifiers trained on frozen features. We report similar results with fine-tuned ResNet-50 models in Figure~\ref{figAppendixWaterbirds}.

\item On \dataset{CivilComments}, we use a standard pretrained BERT. To limit the computational expense for our large number of experiments, we use the BERT-tiny version (2 layers, 2 attention heads, embeddings of size 128).
The results in the main paper use linear classifiers on frozen features.
We report similar results with fine-tuned models in Figure~\ref{figAppendixCivilComments2} (using only one seed).

\item On \dataset{Wild-Time Yearbook}, we train the small CNN architecture described in~\cite{yao2022wild} from scratch.
In the analysis of Figure~\ref{figYearbook2}, we measure the distance between the training and test distributions of inputs (vectorized grayscale images). To do so, we measure the distance between every pair across the two sets.
For each test example, we keep the minimum distance (i.e.\ closest training example), then average these distances over the test set.

\item On \dataset{Wild-Time arXiv}, we use random subset of 10\% of the dataset.
We verified on a small number of experiments that this produces very similar results to the full dataset at a fraction of the computational expense.

\item On \dataset{Wild-Time MIMIC-Readmission},
the baseline transformer architecture proposed in~\cite{yao2022improving} seems inappropriate.
Its ID and OOD performance is surpassed by random guessing or even by constant predictions of the majority training class.
The issue probably went unnoticed because the standard accuracy metric is misleading with imbalanced data ($70\%$ ID accuracy of that ERM baseline is worse than chance).

To remedy this, we first switch to the AUROC metric. It gives equal weight to the classes and $50\%$ is then unambiguously equivalent to random chance.

Second, we use a much simpler architecture. We train a ``bag of embeddings'' where each token (diagnosis/treatment code) is assigned a learned embedding, which are summed across sequences then fed to a linear classifier.
\end{itemize}

All experiments were run on a single laptop with an Nvidia GeForce RTX 3050 Ti GPU.

\section{Additional results}
\label{appendixResults}

We show below results from the main paper while including in-domain (ID), out-of-distribution (OOD) average-domain/average-group, and OOD worst-domain/worst-group performance.
The OOD metrics are always strongly correlated across methods and training epochs, but ID and OOD performance sometimes require a trade-off, as noted recently in~\cite{teney2022id}.

\floatsetup[figure]{margins=centering,capposition=bottom,floatwidth=\textwidth}

\floatsetup[figure]{margins=centering,capposition=bottom,floatwidth=\textwidth}
\begin{figure*}[h!]
  \includegraphics[width=.99\textwidth]{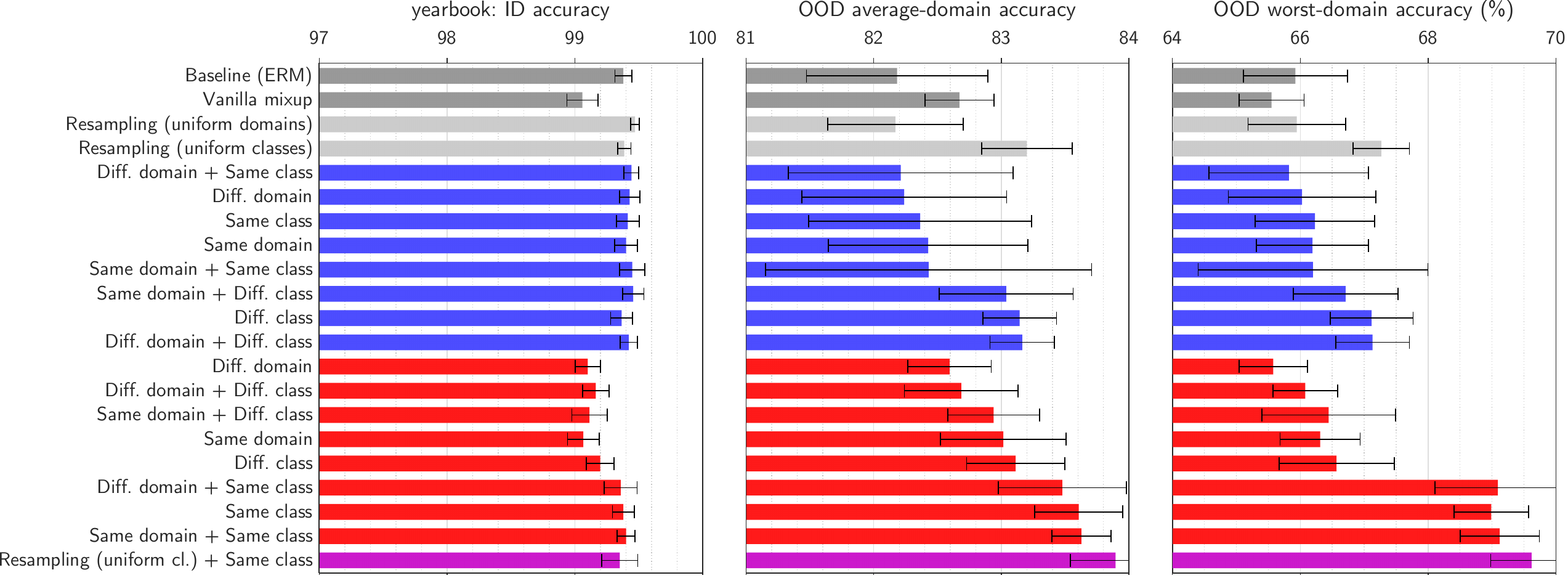}
  \caption{Results on \dataset{yearbook}.
  \label{figAppendixYearbook}}
\end{figure*}

\floatsetup[figure]{margins=centering,capposition=bottom,floatwidth=\textwidth}
\begin{figure*}[h!]
  \includegraphics[width=.99\textwidth]{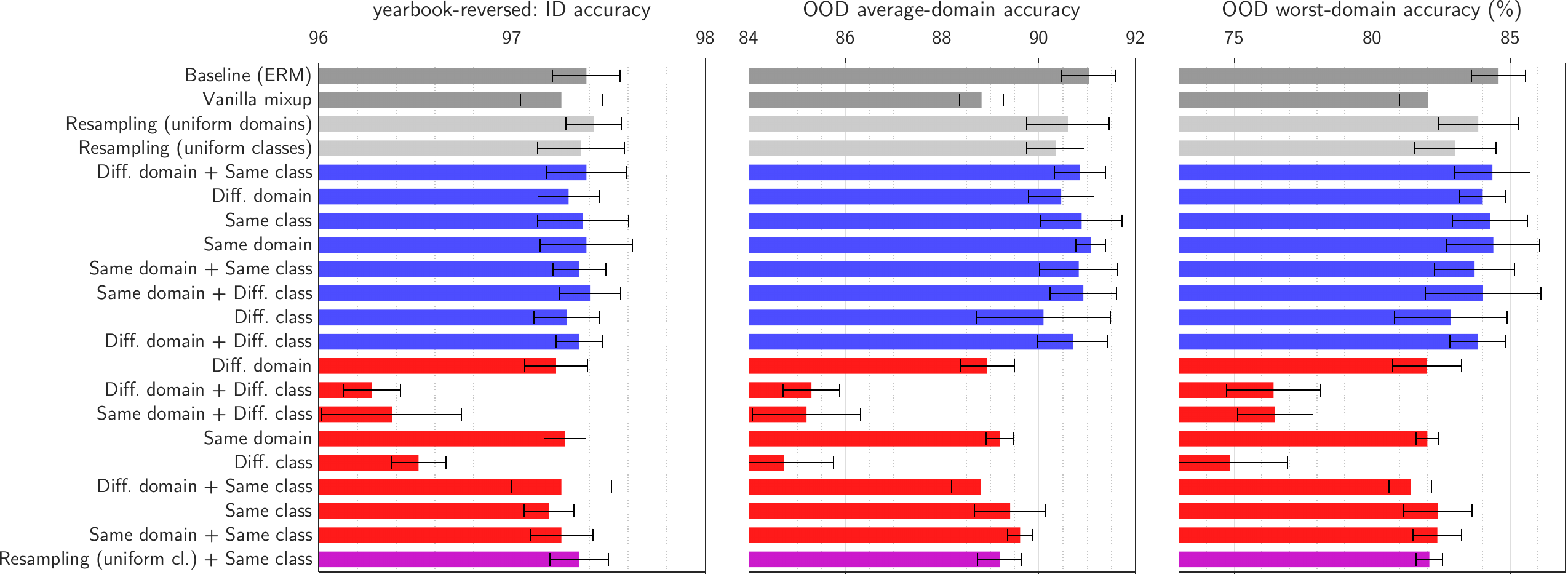}
  \caption{Results on \dataset{yearbook-reversed} (swapping ID and OOD data) to test the predicted failure mode. The ``regression toward the mean'' does not hold, therefore the methods that improved OOD performance on the original dataset are now detrimental (methods presented in the same order as Figure~\ref{figAppendixYearbook}).
  \label{figAppendixYearbookReversed}}
\end{figure*}

\floatsetup[figure]{margins=centering,capposition=bottom,floatwidth=\textwidth}
\begin{figure*}[h!]
  \includegraphics[width=.99\textwidth]{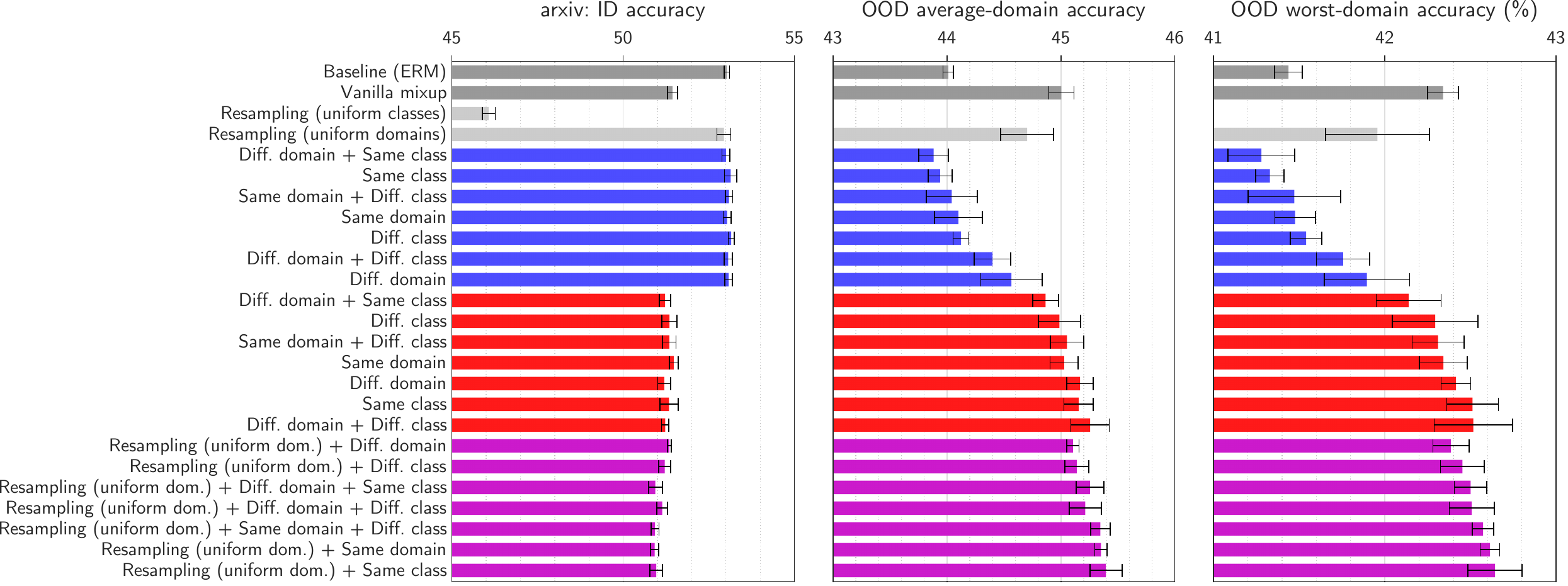}
  \caption{Results on \dataset{arXiv}.
Interestingly, the methods with selective sampling without mixup are much better than selective mixup on in domain (ID) but worse out of domain (OOD). This shows a clear trade-off between these two objectives.
  \label{figAppendixArxiv}}
\end{figure*}

\floatsetup[figure]{margins=centering,capposition=bottom,floatwidth=\textwidth}
\begin{figure*}[h!]
  \includegraphics[width=.16\textwidth]{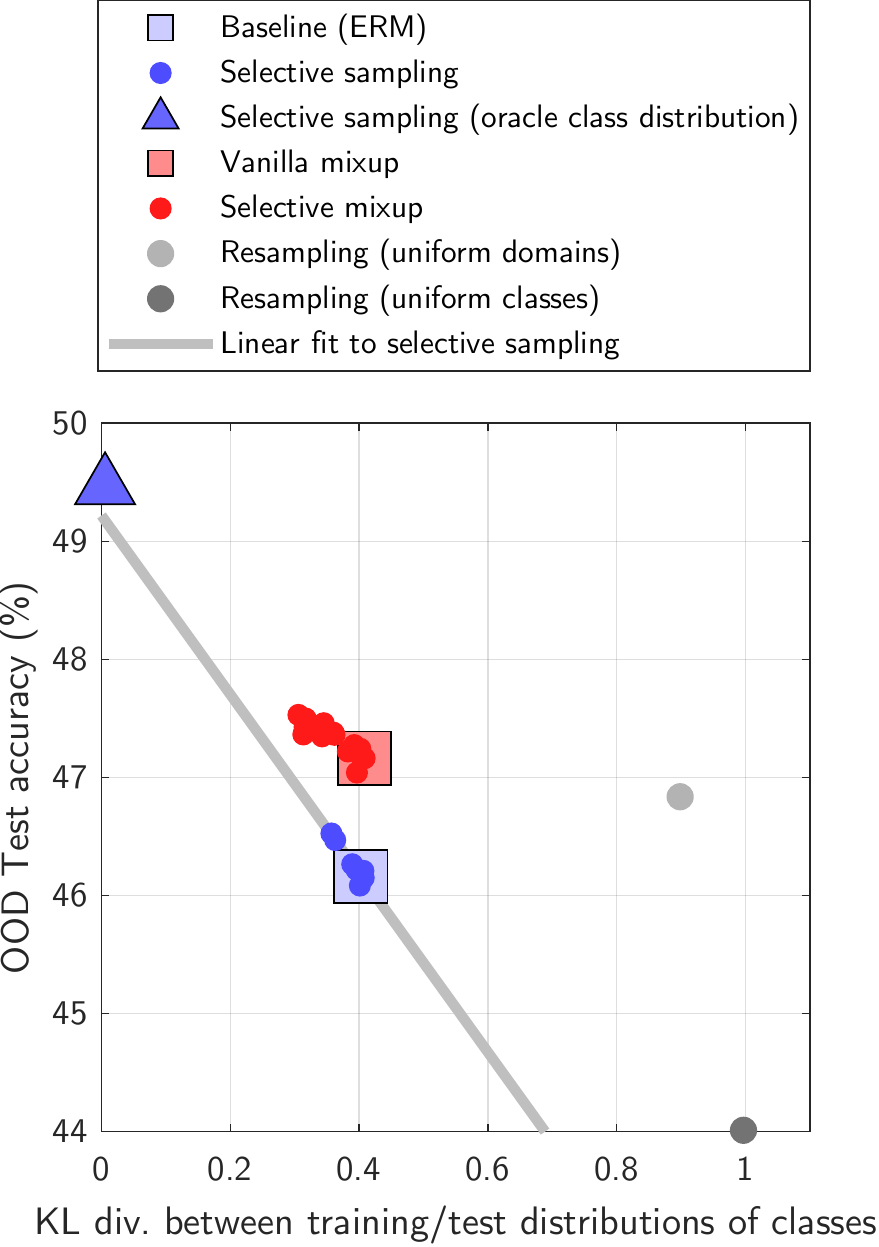}
  \includegraphics[width=.16\textwidth]{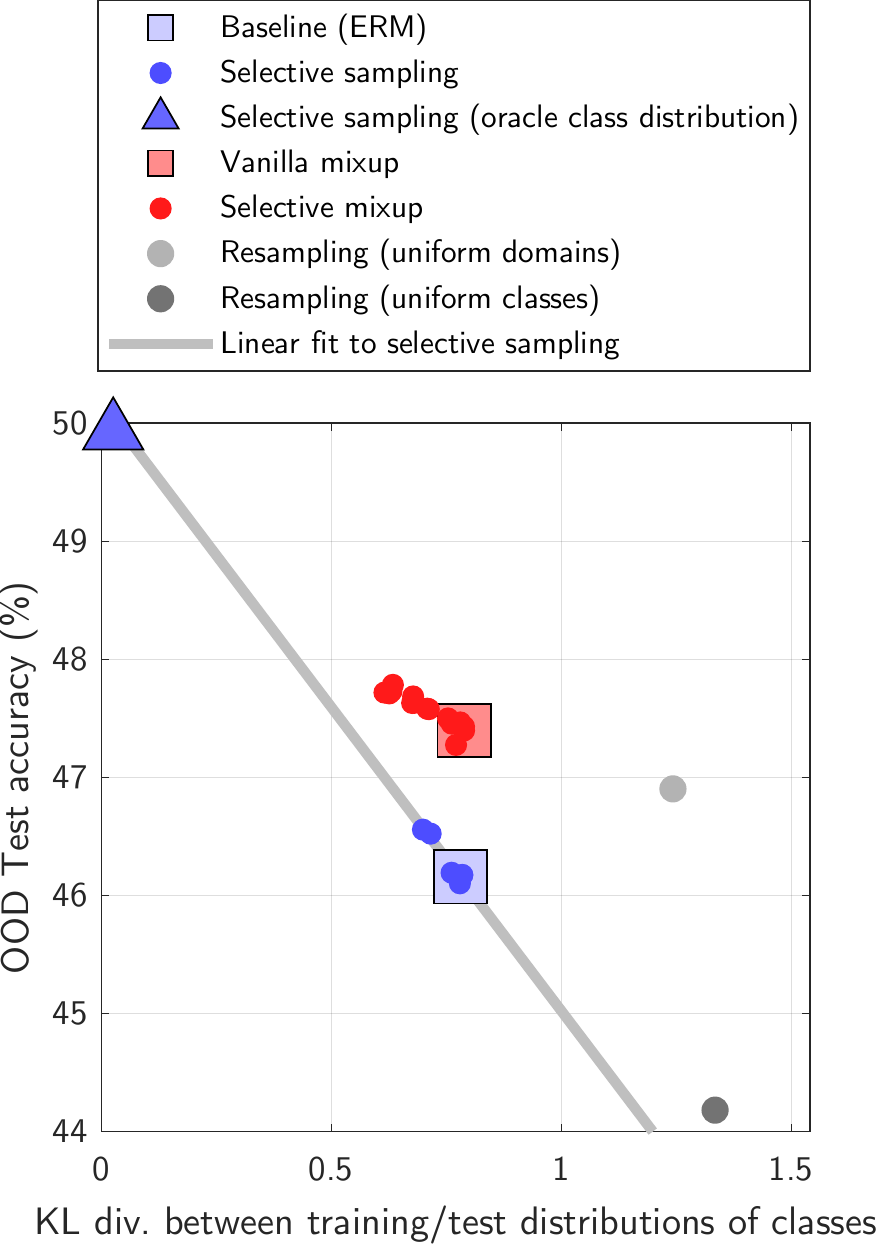}
  \includegraphics[width=.16\textwidth]{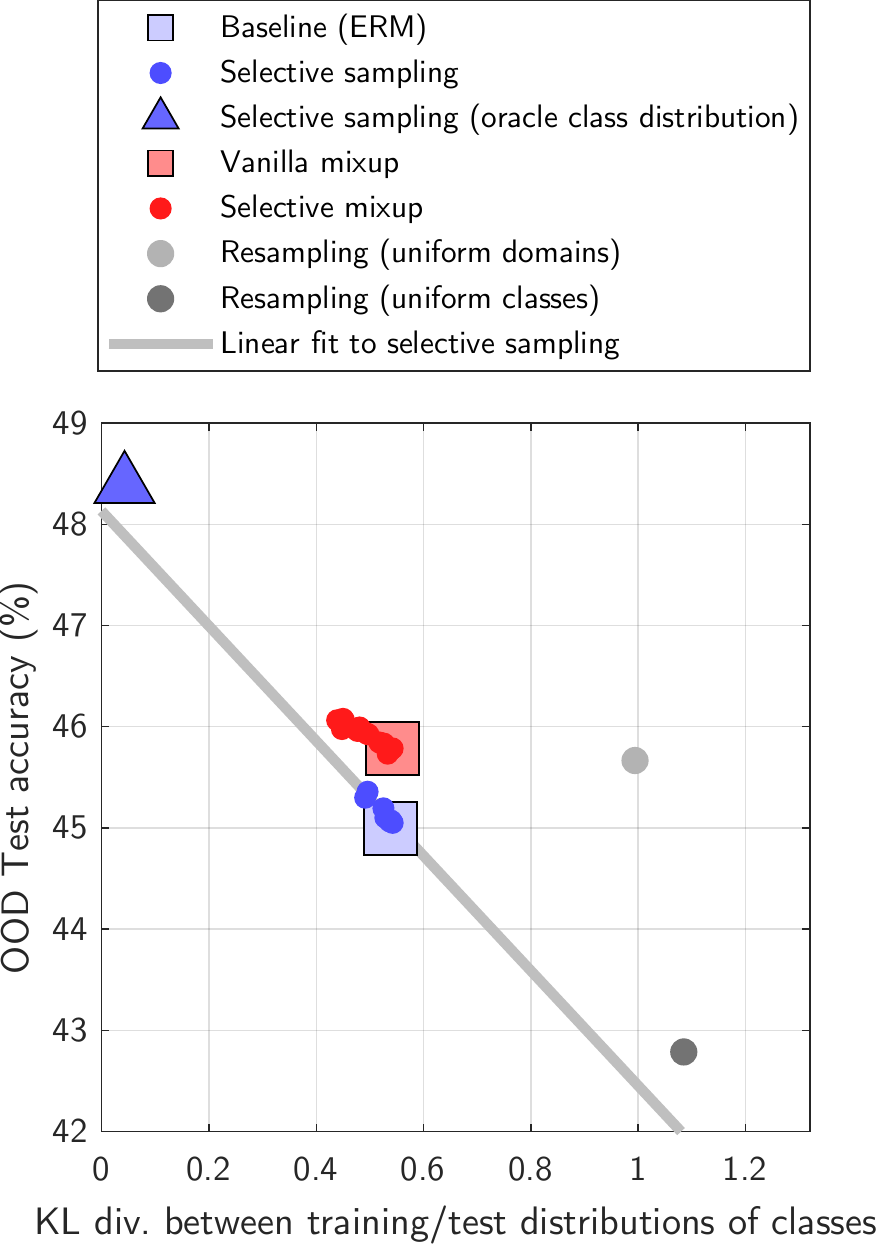}
  \includegraphics[width=.16\textwidth]{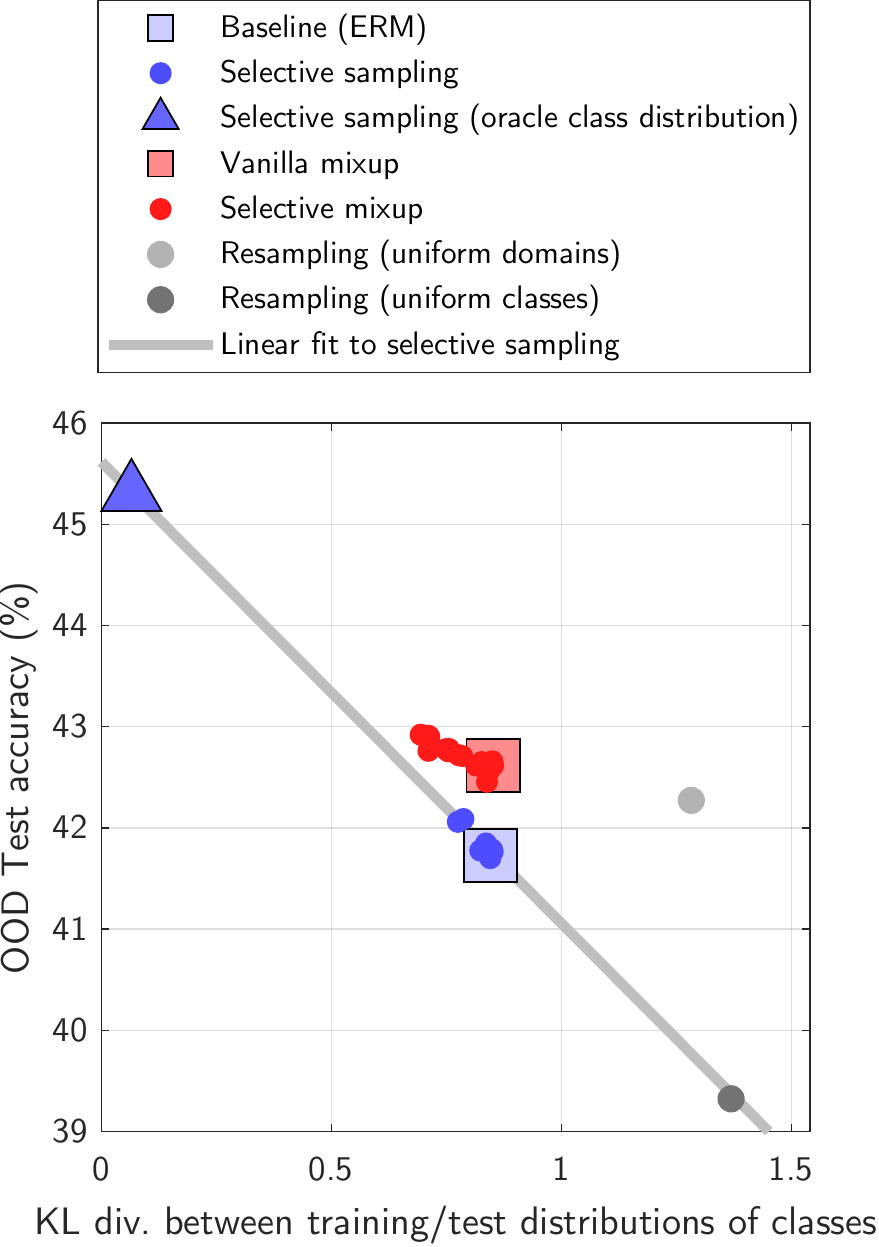}
  \includegraphics[width=.16\textwidth]{figures/arxiv-base-frozen-accVsDist-06.pdf}
  \includegraphics[width=.16\textwidth]{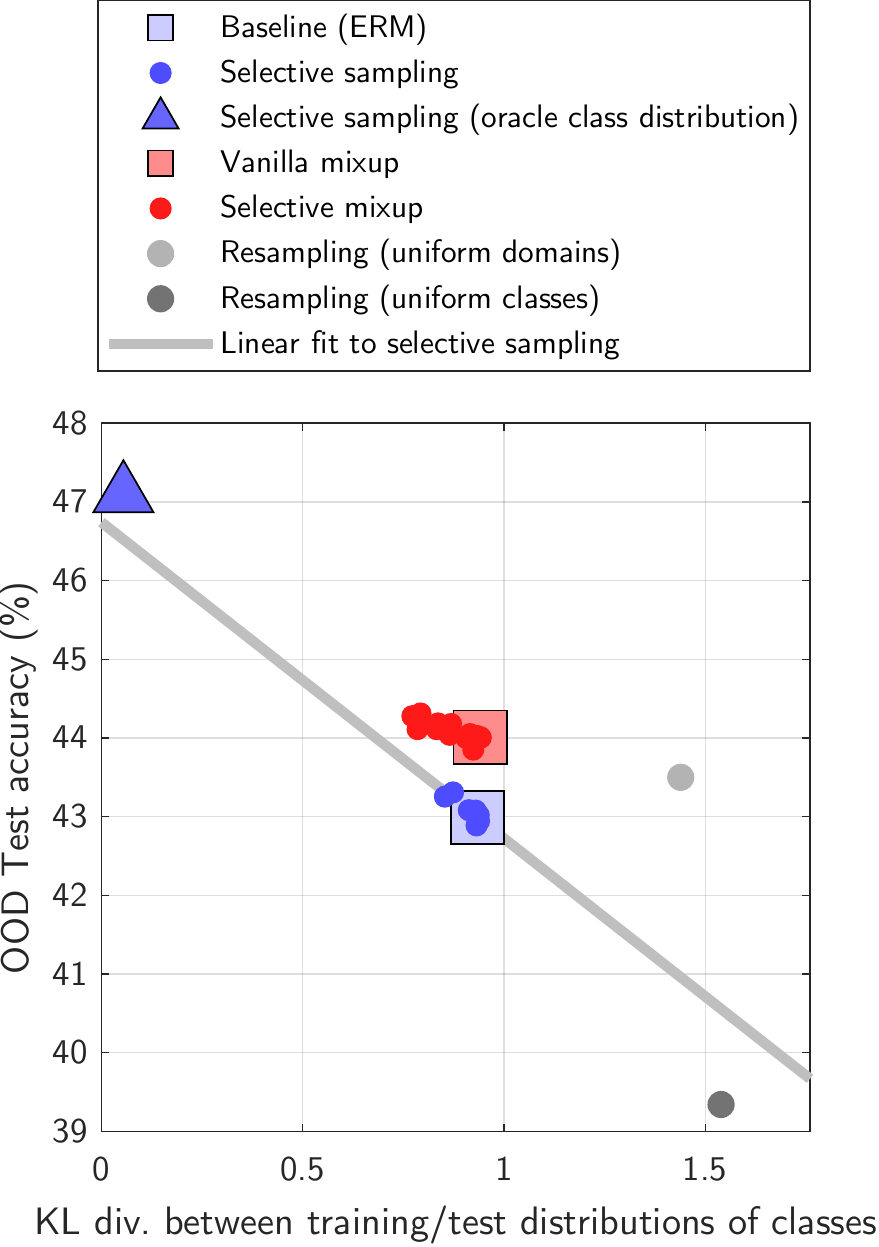}
  \caption{Same analysis as in Figure~\ref{figArxiv2} of the main paper, performed on every test domain.
  In all cases, we observe a strong correlation between the improvements in accuracy and the reduction in divergence of the class distribution due to resampling effects.
  \label{figAppendixArxiv2}}
\end{figure*}

\clearpage
\floatsetup[figure]{margins=hangright,capposition=beside,capbesideposition=right,floatwidth=0.75\textwidth,capbesidewidth=0.23\textwidth}
\begin{figure*}[h!]
  \includegraphics[width=.75\textwidth]{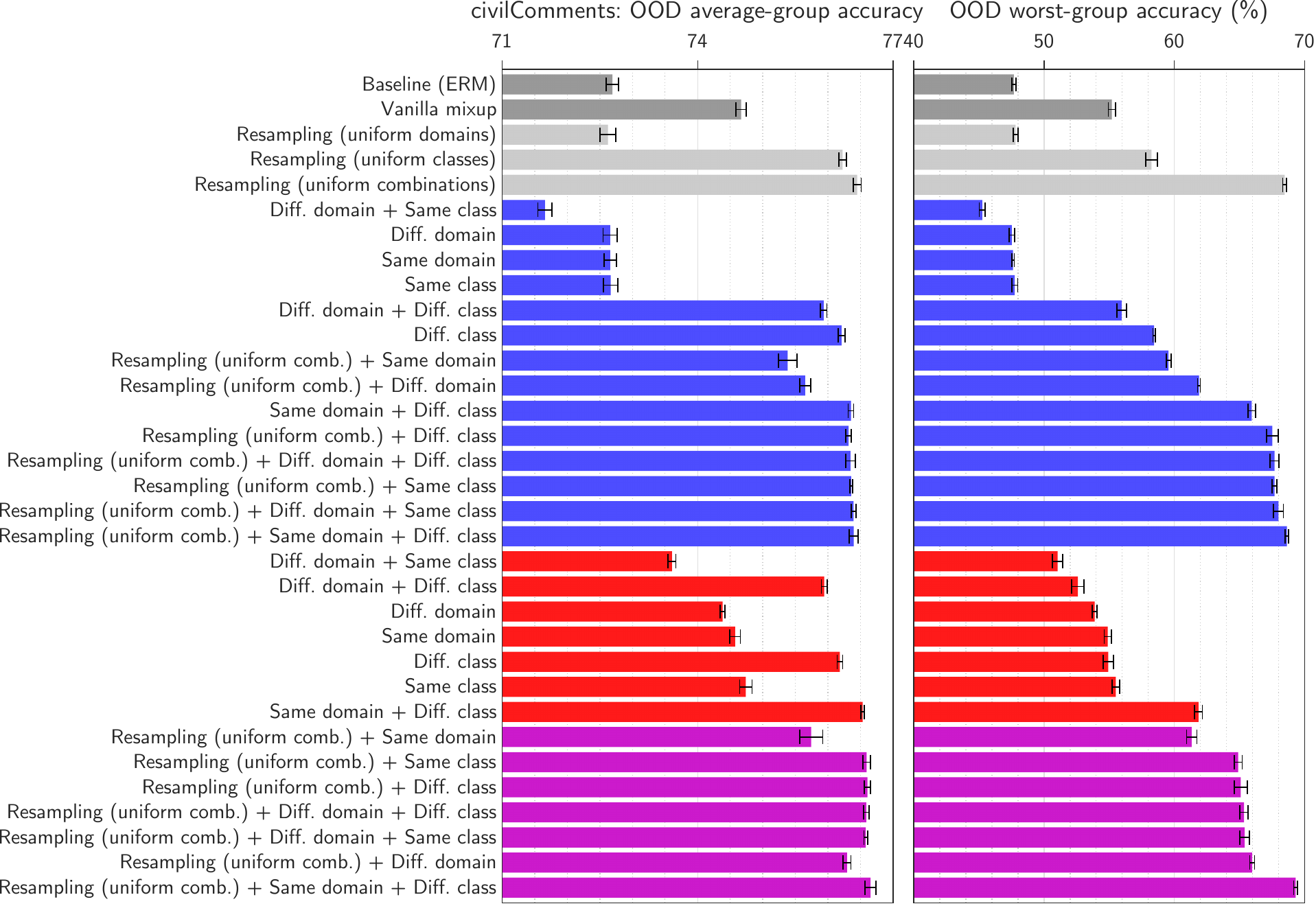}
  \caption{Results on \dataset{civilComments} with linear classifiers on frozen embeddings.
  \label{figAppendixCivilComments}}
\end{figure*}

\begin{figure*}[h!]
  \includegraphics[width=.7\textwidth]{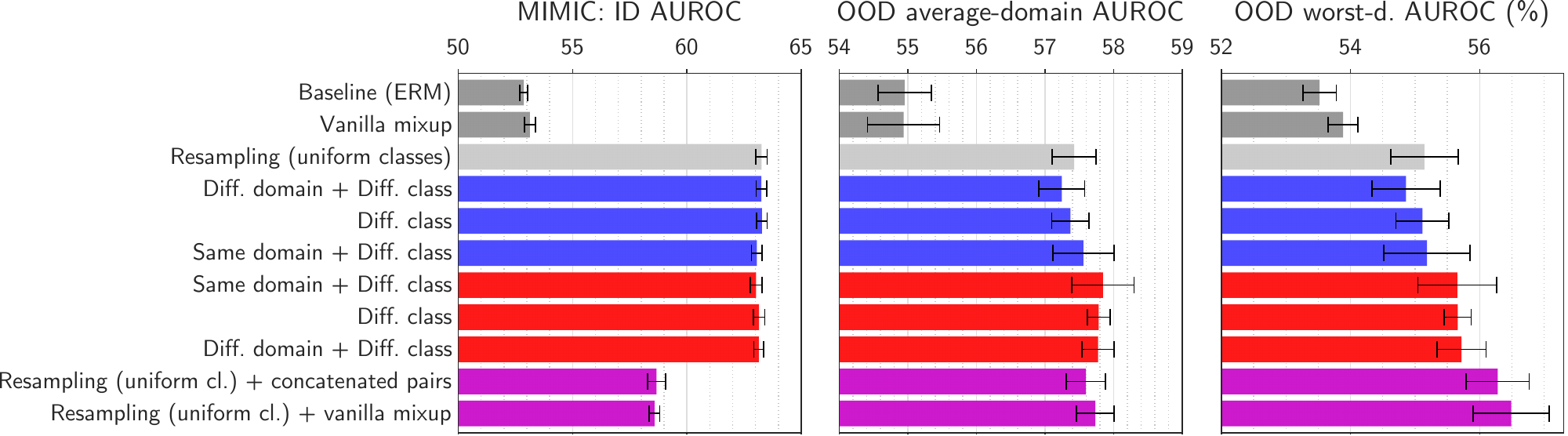}
  \caption{Results on \dataset{MIMIC-Readmission}.
  \label{figAppendixMimic}}
\end{figure*}

\section{Testing the predicted failure mode}
\label{appendixPredictedFailure}

The explanations proposed in this paper state that the resampling effects with selective mixup are beneficial because a ``regression toward the mean'' is present in the datasets.
As a corollary, it implies that the effect would be detrimental if the opposite property holds (i.e.\ increased imbalances in the test data).

We test this prediction on the \dataset{yearbook} dataset by switching the ID and OOD data. More precisely, whereas the original dataset uses data from years 1930--1970 as training data and ID test data (as shown in Figure~\ref{figRegression}), we use this data as the OOD test data. Vice versa for data from years 1970--2010.
As a result, the training\,$\rightarrow$\,test label shift is now an increased imbalance rather than a regression toward a uniform one.

The results on \dataset{yearbook-reversed} confirm the predicted failure (comparing Figures~\ref{figAppendixYearbook}--\ref{figAppendixYearbookReversed}). The methods that improved OOD performance on the original dataset are now detrimental.

\clearpage
\onecolumn
\section{Proof of Theorem~\ref{thmRegression}}
\label{appendixProof}

\abovedisplayskip=7pt
\belowdisplayskip=7pt

\begin{theorem}[Restating Theorem~\ref{thmRegression}]
    Given a dataset
    $\mathcal{D} \!=\! \{(\bx_i,\by_i)\}_i$
    and paired data $\widetilde{\mathcal{D}}$
    sampled 
    according to the ``\crit{different class}'' criterion, i.e.\
    $\widetilde{\mathcal{D}} = \{ (\widetilde{\bx}_i, \widetilde{\by}_i) \sim \mathcal{D}
    ~~\textrm{s.t.}~~
    \widetilde{\by}_i\!\neq\!\by_i\}$,
    then the distribution of classes in
    $\mathcal{D} \mkern-0.6mu\cup\mkern-0.6mu \widetilde{\mathcal{D}}$
    is more uniform than in
    $\mathcal{D}$.
    
    Formally, the entropy\;\,
    $\entropyBig{\pY(\mathcal{D})} \,\le\, \entropyBig{\pY(\mathcal{D} \cup \widetilde{\mathcal{D}})}$.

\begin{proof}
Let us define the shorthands
$\bp \eqdef \pY(\mathcal{D})$
~and~
$\bpp \eqdef \pY(\widetilde{\mathcal{D}})$.

In $\widetilde{\mathcal{D}}$, the $i$th class gets assigned, in the expectation, on a proportion of points equal to the proportion of all other classes $j \neq i$ in the original data $\mathcal{D}$.

Looking at the individual elements of $\bpp$, we therefore have, $\forall\, i\!=\!1\ldots C$:
\begin{align}\label{eqProof}
    \bppi &~=~ \Sigma_{i \neq j}^C \,\bpj &\;/\; (C\!-\!1)\\
    \bppi &~=~ (1 \!-\! \bpi) &\;/\; (C\!-\!1)
\end{align}
We will show that every element of $\bpp$ is closer to $\tfrac{1}{C}$
than the corresponding element of $\bp$:
\begin{align}
    |\bpi - \tfrac{1}{C}| &~\ge~ |\bppi - \tfrac{1}{C}|\\
    | \tfrac{C \, \bpi - 1}{C}| &~\ge~ | \tfrac{ (1-\bpi) \, C - (C - 1)}{C \,(C-1)} |\\
    | C \, \bpi - 1| &~\ge~ | \tfrac{ 1 \,-\, C \, \bpi}{(C-1) } |\\
    | C \, \bpi - 1| &~\ge~ | \tfrac{ C \, \bpi \,-\, 1}{(C-1) } |
\end{align}
Therefore $\bpp$ is closer to a uniform distribution than $\bp$, and 
\begin{align}
    \entropy{\bp} &~\le~ \entropy{\bpp}
\end{align}
Since 
$\pY(\mathcal{D} \cup \widetilde{\mathcal{D}}) = \big( \pY(\mathcal{D}) \,\oplus\, \pY(\widetilde{\mathcal{D}})\big) \,/\, 2$,
we also have
\begin{align}
\entropyBig{\bp} &~\le~ \entropyBig{ (\bp \,\oplus\, \bpp) / 2 }\\
\entropyBig{\pY(\mathcal{D})} &~\le~ \entropyBig{\pY(\mathcal{D} \cup \widetilde{\mathcal{D}})}
\end{align}
with an equality iff $\pY(\mathcal{D})$ is uniform. 
\end{proof}
\end{theorem}


\end{document}